\documentclass[journal]{IEEEtran}
\usepackage{cite}
\usepackage{color}
\usepackage{subfigure}
\ifCLASSINFOpdf
   \usepackage[pdftex]{graphicx}
\else
\fi
\usepackage{amsmath}
\usepackage{bm}
\usepackage{amssymb}
\usepackage{amsthm}
 \newenvironment{sistema}%
  {\left\lbrace\begin{array}{@{}l@{}}}%
  {\end{array}\right.}

\usepackage{algorithm}
\usepackage{algorithmic}

\usepackage{array}
\usepackage{url}

\usepackage{color}
\usepackage{soul}
\definecolor{orange}{rgb}{0.93, 0.53, 0.18}

\newcommand{\LorenzoDelete}[1]{{\st{#1}}}
\newcommand{\LorenzoAdd}[1]{{\color{black}#1}}
\newtheorem{remark}{Remark}
\definecolor{ao}{rgb}{0.0, 0.5, 0.0}
\newcommand{\GiuliaDelete}[1]{}
\newcommand{\GiuliaAdd}[1]{{\color{black}#1}}

\begin{document}
%
\title{Memory 
Unscented Particle Filter for \\ 6-DOF Tactile Localization}
%
%
%

\author{G.~Vezzani, 
            U.~Pattacini, 
            G.~Battistelli, L.~Chisci, and L.~Natale
\thanks{L. Natale (lorenzo.natale@iit.it), U. Pattacini (ugo.pattacini@iit.it) and G. Vezzani (giulia.vezzani@iit.it) are with the Italian Institute of Technology (IIT), iCub Facility, Via Morego 30, Genova, Italy. 
G. Vezzani is also with the University of Genova, Via All'Opera Pia, 13, 16145 Genova. 
G. Battistelli (giorgio.battistelli@unifi.it) and L. Chisci (luigi.chisci@unifi.it) are with the Dipartimento di Ingegneria dell'Informazione, Universit\`a degli Studi di Firenze, Via S. Marta 3, Firenze, Italy.
}
\thanks{}
\thanks{}}

\maketitle

\begin{abstract}
This paper addresses $6$-DOF (degree-of-freedom) tactile localization, i.e. the pose estimation of  tridimensional objects given tactile measurements. This estimation problem is fundamental  for the operation of autonomous robots that are often required to manipulate and grasp objects whose pose is a-priori unknown. The
nature of tactile measurements, the strict time requirements for real-time operation and the multimodality of the involved probability distributions pose remarkable challenges and call for advanced nonlinear filtering techniques. Following a Bayesian approach, this paper proposes a novel and effective algorithm, named \textit{Memory Unscented Particle Filter} (MUPF), which solves the $6$-DOF localization problem recursively in real-time by only exploiting \textit{contact point} measurements.
MUPF combines a modified particle filter that incorporates a sliding memory of past measurements to better handle multimodal distributions, along with the unscented Kalman filter that moves the particles towards regions of the search space that are more likely with the measurements.
The performance of the proposed MUPF algorithm has been assessed both in simulation and on a real robotic system equipped with tactile sensors (i.e., the iCub humanoid robot). The experiments show that the algorithm provides accurate and reliable localization even with a low number of particles and, hence, is compatible with  real-time requirements.

\end{abstract}


%
\IEEEpeerreviewmaketitle

\section{Introduction}
%
%
%
%
\IEEEPARstart{A}{ccurate} perception is a necessary requirement for general operation of autonomous robots in real-world environments, and, in particular, for object manipulation.
The development of new sensors and inference techniques plays a major role in improving manipulation capabilities: it enhances the robot ability to deal with uncertainties,  increases flexibility and also reduces the cost required to engineer the environment in which the robot will operate.
In robotics, the use of vision has been thoroughly investigated \cite{visionrew}, while the interest in tactile sensors is quite recent.
This is due to the fact that only recent advances in tactile technology have made it possible to build tactile systems that are reliable enough and can be deployed on real robots at a reasonable price \cite{fingertip},\cite{methods}, \cite{biomimetic}.
 This  recent improvement of tactile sensors is one of the reasons for a surge of interest on this topic \cite{tactrew}. 
Findings in human physiology testify how humans jointly exploit vision and touch in order to accomplish manipulation tasks and how humans are even able to explore objects by means of tactile perception solely \cite{phys1}, \cite{phys2}.  Thus, improvements  on the use of touch sensors will make manipulation tasks more efficient, complementing vision when unavailable or imprecise, for example due to occlusions and/or bad lighting conditions. 
\\
\indent Due to technological limitations, tactile systems have low resolution and rarely provide other than estimation of the force normal to the surface. 
Object localization using tactile feedback is, therefore, challenging and requires the development of filtering techniques that allow appropriate fusion of multiple measurements, taking into account the presence of noise and the real-time requirements of the task.

 \indent This paper proposes  a novel algorithm for the solution of tridimensional object tactile localization, named \textit{Memory Unscented Particle Filter }(MUPF). This algorithm relies on the  \textit{Unscented Particle Filter} (UPF) \cite{upf}, conveniently modified so as to efficiently solve the global 6-DOF (degree-of-freedom) localization problem, by exploiting contact point measurements only. 
The proposed MUPF algorithm has proved to be effective in experimental validation carried out in simulation as well as in a real setting using the iCub humanoid robot and its tactile system~\cite{fingertip}.\\
 \indent The paper is organized as follows. 
 Section \ref{rel} provides an overview on tactile localization. 
 Section \ref{math}  is a brief introduction to nonlinear  filtering  techniques useful for the subsequent developments of the paper. 
 Section \ref{prob} provides  a mathematical (Bayesian) formulation of the tactile localization problem. 
 Section \ref{mupfsec} presents the novel \textit{Memory Unscented Particle Filter} (MUPF) approach to 6-DOF tactile localization. 
 Section \ref{sim} and \ref{exp} demonstrate the effectiveness of the proposed approach by means of simulation and, respectively, experimental tests on the iCub humanoid robot. 
 Finally, Sections  \ref{disc} and \ref{conc} end the paper with concluding remarks and perspectives for future work.

\section{Related Work}\label{rel}

\indent The first contributions on tactile object localization (1980s) tackled the problem by using mostly  iterative optimization methods  and 
focused on finding a single solution best fitting the set of available measurements \cite{gaston,grimson,faugeras}. 
Since these methods tend to be trapped in local minima, \emph{low initial uncertainty} is assumed so as to ensure that the optimization algorithm is initialized near the solution. 
In order to avoid local minima, the algorithm can be executed multiple times from different starting points.\\
\indent Over the last years, Bayesian methods have been playing an important role in tactile localization  \cite{markov, chap, negative, Petr}. 
In particular, these methods  are capable of working with noisy sensors, inaccurate models, moving objects and can give information on \emph{where to sense next} during tactile exploration. \normalfont Thus, they can be used not only to localize the object, but also to provide useful information for collecting measurements and  for real exploration. \\
 \indent Since the localization problem is intrinsically of multimodal nature (i.e. the probability density exhibits multiple peaks), nonlinear Kalman filtering techniques (such as the extended or unscented Kalman Filter) cannot be satisfactorily used. 
 In this respect, the Bayesian framework (e.g. particle filtering) is more appropriate, since it intrinsically handles multimodal distributions. 
 On the other hand, its main drawback is represented by the computational complexity, which grows exponentially  with the number of independent variables (DOFs) and polynomially with the size of the initial region of uncertainty. 
 For example, recalling that the localization of an object involves 6 DOFs, a particle filter should be configured to run with a number of particles in the order of 10$^6$, which might entail an unaffordable computational burden for real-time operation. 
In fact, most of the existing work is characterized by assumptions limiting 
the number of DOFs and the size of initial uncertainty.\\ \indent 
In this respect, the first known work  traces back to 2001 and is due to \slshape Gadeyne et al.\normalfont \cite{markov},   who performed 3-DOF localization of a box  with initial uncertainty of $300 ~mm$ in position and $360$ degrees in orientation. 
Measurements were taken by a force-controlled robot and a sampled measurement model, stored in a lookup table and constructed off-line, was used. 
In 2005, \slshape Chhatpar et al.\normalfont \cite{chap} used particle filters to achieve 3-DOF localization with $20~ mm$ of initial uncertainty in peg-in-hole assembly tasks. The exploited measurement model was obtained by sampling the object in advance. \\\indent 
An interesting approach to tactile localization makes use of the \textit{Scaling Series} method \cite{Petr2,Petr} developed by \slshape Petrovskaya et al.\normalfont, by which  6-DOF localization  has been achieved with  large initial uncertainty of 
$400~ mm$ in position and $360$ degrees in orientation. 
This method, which combines Bayesian Monte Carlo and annealing techniques, exploits measurements  of contact points and surface normals.  
It performs multiple iterations over the data, gradually scaling precision from low to high. 
For each iteration, the number of particles is automatically selected on the basis of the complexity of the annealed posterior. \\ 
\indent In 2010, \slshape Corcoran et al.\normalfont \cite{negative} used an annealed particle filter to estimate a 4-DOF pose and radius of cylindrical objects. 
The initial uncertainty was of $250~ mm$ in position and unrestricted in orientation. 
The measurement model proposed in \cite{Petr} was extended by exploiting the concept of ``negative information''.
To this end, a set of ``no-contact measurements'' is defined to account for regions explored by the robot 
where it is known or it can be inferred that the object cannot be located, since no contacts are perceived. \\
 \indent In 2013, \slshape Koval et al. \normalfont  \cite{manifold} addressed object localization  during manipulation actions which involve persistent contact (e.g. pushing) and proposed a modified particle filter to estimate the state of the object. 
 In the same year, \slshape Chalon et al. \normalfont \cite{online} presented another particle filter method including information on both object and finger movements. 
 A recent work  \cite{bimbo} combines global optimization methods with the Monte Carlo approach in order to provide a batch solution to the global localization problem, either improving the estimate of the object pose obtained by vision or globally estimating pose when vision is not available.\\
 \GiuliaDelete{\indent  In order to fully exploit touch sensors, a  tactile exploration strategy must be defined: tactile measurements for localization  must be collected by robots autonomously and efficiently. 
A lot of recent work has focused on such a goal. 
In 2010, \slshape Hsiao et al. \normalfont \cite{hsiao} developed a general-purpose strategy for task-driven manipulation of objects. 
They used grids for the 3-DOF pose estimation and proposed optimized data-collection strategies, considering free-standing objects. 
Another noteworthy contribution has been  provided by \slshape{Hebert et al.}\normalfont \cite{nextbest} (2013), in which a tactile method based on  an information gain metric allows a robot, equipped with suitable sensors, to choose the next sensing action so as to accurately localize the object of interest. The method was also extended to choose the most informative action to simultaneously localize and estimate the object's model parameters or model class, but the attention is concentrated on the exploration strategy rather than on the estimation algorithm.
The problem of \emph{where to sense next} is faced  also in \cite{actionsel}, that presents a scheme for computing the next best action so as to optimize a suitable trade-off between an  information gain  and the time required for computation and motion execution. \\} 
 \indent From a technological standpoint, the most commonly used sensors  for tactile localization  are force sensors, usually located on the end effector of industrial manipulators, which provide both contact point and surface normal measurements \cite{markov,Petr, actionsel}. 
 Conversely, more recent works - such as {\cite{negative,manifold,online,nextbest} and the present paper - 
rely on a human-like robot hand   to retrieve measurements, with a different kind of tactile sensors. 
An example of capacitive tactile sensors located on the fingertips of robot hands are the ones described in \cite{fingertip}, able to retrieve the pressure exerted on the fingertip when contact is detected. 
\GiuliaDelete{In this respect, a recent paper  \cite{lepora} \normalfont provides a 
   detailed and systematic analysis of tactile superresolution (SR) localization  for a biomimetic fingertip and a flat region of tactile skin for humanoid robots. 
In such a work,  it is shown that the tactile SR is of paramount importance for the design and application of artificial tactile sensors. 
Three key factors  enable the perceptual acuity to surpass the sensor resolution:  
(1) sensors constructed with multiple overlapping,
broad but sensitive receptive fields;
(2)  the use of the tactile perception method interpolating
   between receptors (taxels) to attain subtaxel acuity;  (3)
   the active perception ensuring robustness to the unknown initial contact
   location. }\\ 
\indent This paper proposes a novel algorithm, the \emph{Memory Unscented Particle Filter}, to solve the $6$-DOF object localization using tactile measurements. 
The algorithm was designed to exploit only the measured position of the contact points obtained from the tactile sensors on the robot unlike other work in  the literature,
 wherein available measurements also include  the surface normal at the contact points \cite{Petr,hsiao,actionsel} or consist of a  6-dimensional vector including  force and torque.~\cite{markov}. 

\textcolor{black}{It is worth pointing out that the proposed solution is inherently recursive in that
the measurements are sequentially processed  in real time as they become available, and the algorithm can provide the object's pose estimate after the processing of each measurement.
There are several reasons for considering a recursive approach: 
 the algorithm can provide the object's pose estimate after the processing of each measurement, and not only at the final measurement acquisition time as with a batch procedure like the one in \cite{Petr2,Petr};
 it is compliant with active exploration techniques where the robot decides, at each time $t$, where to sense next on the basis of the current object's pose estimate;
 it can allow stopping the object localization procedure at a given time $t$ whenever a suitable stopping criterion is satisfied.
 }

The main novelties of this work are: 
\begin{enumerate}
\item the adoption of the \emph{Unscented Particle Filter} (UPF) \cite{upf}, i.e. a variant of the particle filter exploiting an UKF for each particle, for tactile object localization;
\item the definition of an appropriate measurement model for the touch sensor necessary for the application of UPF to tactile-based localization; 
\item a suitable modification of UPF to enforce \textit{memory} of past measurements in the update of the importance  weights.
\end{enumerate} 
Experiments in simulations as well as with a real robotic platform show that, with respect to the state-of-the-art, MUPF is significantly more reliable, requires much less parameter tuning and is able to localize the object with comparable precision.

\section{Mathematical Background}\label{math}

Hereafter, \textit{tactile localization} is cast into the Bayesian framework and addressed as a nonlinear multimodal filtering problem. 
Recall that filtering is the problem of recursively estimating the state \textcolor{black}{$\bm{x}_t \in \mathbb R^n$} of a dynamical system while acquiring and processing noisy observations on-line. 
Specifically, from a Bayesian viewpoint, the goal of the filtering problem is to recursively compute the following conditional PDFs
\begin{equation}
\begin{split}
\text{ } \text{ }p_{t|t}(\bm{x})& = p(\bm{x}_t = \bm{x} | \bm{y}^t)\\
p_{t+1|t}(\bm{x})&= p( \bm{x}_{t+1} = \bm{x} | \bm{y}^{t}),
\end{split}
\end{equation}
given the noisy observations $ \bm{y}^t=\{\bm{y}_1, \dots, \bm{y}_t \}$ \textcolor{black}{with $\bm{y}_t \in \mathbb R^p$}.



The solution of the filtering problem is given by the Bayesian recursion, starting from the initial prior $p_{1|0} (\cdot)$ and consisting of
two functional equations, i.e. the following Bayes and respectively Chapman-Kolmogorov equations:
 \begin{equation}
 p_{t|t}(\bm{x})= \frac{\ell_t(\bm{y}_t|\bm{x})p_{t|t-1}(\bm{x})}{\int \ell_t(\bm{y}_t| \bm{\xi}) p_{t|t-1}(\bm{\xi})d\bm{\xi}}
 \label{bayes}
 \end{equation}
 \begin{equation}
 p_{t+1|t}(\bm{x})= \int \varphi_{t+1|t}(\bm{x}| \bm{\xi}) p_{t|t}(\bm{\xi})d\bm{\xi}
 \label{chapman}
 \end{equation}
\textcolor{black}{where $ \varphi_{t+1|t}(\bm{x}|\bm{\xi})$ is the {\em Markov transition density} representing the conditional probability that the state at time $t+1$ will take value $\bm{x}$ given that the state at time $t$ 
is equal to $\bm{\xi}$, and $\ell_{t} (\bm{y} | \bm{x}) $ is the {\em measurement likelihood function} denoting the probability that the measurement at time $t$ will take value $\bm{y}$ given that the state is equal to $\bm{x}$.
 }

However, in many practical applications,  such as navigation, tracking and localization, 
the transition and likelihood models are usually affected by nonlinearities and/or non-Gaussian noise distributions, thus precluding analytical solutions
\textcolor{black}{of \eqref{bayes} and \eqref{chapman}}. In these cases,
one must invariably resort to some approximation technique.
Most of the existing approximation techniques can be divided in two families: Kalman-filtering-like approaches, and sequential Monte Carlo methods. 
The algorithms belonging to the former family (like the {\em Extended Kalman filter} and the {\em Unscented Kalman filter} \cite{julieruhlmann1996}-\cite{julieruhlmann1997}) propagate only the first- and second-order moments 
(i.e., mean and covariance) of the posterior state distribution. Such methods are usually characterized by a low computational cost, 
but are not appropriate for multimodal distributions like the one arising from the tactile localization problem.
On the other hand, sequential Monte Carlo methods,  also known as \emph{particle filters} \cite{doucet}, can deal with arbitrary nonlinearities and distributions and supply 
a complete representation of the posterior state distributions.

 Particle filtering techniques stem from the idea of approximating the posterior density function $p_{t|t}(\bm{x})$ by means of a finite set of weighted samples (particles) as
 \begin{equation}\label{PF}
  \hat{p}_{t|t}(\bm{x}) \approx \sum_{i=1}^N{\tilde {w}_t^{(i)} \text{ }\delta(\bm{x}- \bm{x}_{t|t}^{(i)})}, 
 \end{equation}
 where $\delta (\cdot)$ is the Dirac delta function,  $\bm{x}_{t|t}^{(i)}$ is the position of the $i$-th particle and  $\tilde w_{t}^{(i)}$ its normalized importance weight.
 In this way, the evaluation of the integrals that are necessary for application of the Bayesian filtering equations \eqref{bayes} and \eqref{chapman} is performed via the Monte Carlo numerical integration method,
 i.e., by transforming the integrals into discrete sums.
 
In principle, the particle approximation (\ref{PF}) can be computed by drawing a set of independent and identically distributed samples $\{\bm{x}_{t|t}^{(i)}, i=1, \dots, N\}$ from the posterior $p_{t|t}(\bm{x})$.
While such a solution is not feasible because $p_{t|t}(\bm{x})$ is not known, the difficulty can be circumvented by sampling each particle $i$ from a known, easy-to-sample, \emph{proposal distribution} $q^{(i)} (\bm{x}_t | \bm{y}^t)$,
and then compute the normalized importance weights as
\begin{eqnarray}\label{wrec}
  w_t^{(i)} &=& \tilde w_{t-1}^{(i)} \frac{\ell_t( \bm{y}_t|\bm{x}_{t|t}^{(i)})\text{ }\varphi_{t|t-1}(\bm{x}_{t|t}^{(i)} |\bm{x}_{t|t-1}^{(i)})}{q^{(i)} (\bm{x}^{(i)}_{t|t} | \bm{y}^t)} \, , \\
  \tilde w_t^{(i)} &=& w_t^{(i)} / \sum_{j=1}^N   w_t^{(j)} \, .
\end{eqnarray}
In fact, by comparing (\ref{wrec}) with \eqref{bayes} and \eqref{chapman}, it is an easy matter to see that the resulting particle-based description approximates the true posterior $p_{t|t}(\bm{x})$ at time $t$.

 \subsection{The Unscented Particle Filter}

\textcolor{black}{The main drawback of particle filtering techniques is that, unless special care is taken,
the number $N$ of particles needed to make the approximation (\ref{PF}) sufficiently accurate can 
 increase exponentially with the dimension
 $n$ of the vector to be estimated (since it is required to sample in a subset of $\mathbb R^n$).
In this respect, a critical point of particle filtering is how to choose the 
proposal distribution $q^{(i)} (\bm{x}_t | \bm{y}^t)$ so as to
approximate the posterior reasonably well with a moderate number of particles. }
Among the most effective variations, there is the  \emph{unscented particle filter} (UPF) which exploits the UKF \textcolor{black}{in the proposal distribution} to improve performance \cite{upf}. In the following part of this section, 
an outline of the UPF algorithm  is provided.

The UPF propagates a set
 of extended particles $\mathcal P_t = \{ \mathcal P_t^{(1)}, \ldots,  \mathcal P_t^{(N)} \}$, each one comprising a weight $\tilde {w}_t^{(i)}$, a mean $\bm{x}_{t|t}^{(i)}$, and a covariance $P_{t|t}^{(i)}$, i.e.,
 \[
 \mathcal P_t^{(i)} = \{ \tilde {w}_t^{(i)} , \bm{x}_{t|t}^{(i)} , P_{t|t}^{(i)} \} \, .
 \]
\textcolor{black}{Given the set of particles at time $t-1$, the UKF prediction and correction steps are 
applied to each particle mean and covariance so as to move the particle towards the measurements. 
Then, for each $i$, a new particle is sampled using $\mathcal{N} (\bm{x}_t;\bar{\bm{x}}_t^{(i)}, P_{t|t}^{(i)})$ as proposal distribution
where $\bar{\bm{x}}_t^{(i)}$ is the updated mean after the correction step, $P_{t|t}^{(i)}$ is the updated covariance, and $\mathcal N(\bm{x};\bar{\bm{x}},P)$ denotes the normal distribution with mean $\bar{\bm{x}}$ and covariance $P$,
}
thus achieving a more dense sampling in the most relevant areas
of the search space.

In order to apply the UKF to each particle, it is necessary to assume that the Markov transition density $\varphi_{t+1|t} (\bm{x}_{t+1}| \bm{x}_{t})$ and measurement likelihood function 
$\ell _t(\bm{y}_t | \bm{x}_t )$ are generated by a state transition and, respectively, measurement equation,
so that the time evolution of $\bm{x}_t$ and $\bm{y}_t$ can be described by 
the discrete-time dynamical system 
\begin{eqnarray}\label{dynsyst1}
\bm{x}_{t+1} &=&\bm{f}_t(\bm{x}_t,\bm{\omega}_t) \\
\bm{y}_t &=& \bm{h}_t(\bm{x}_t,\bm{\nu}_t) . \label{dynsyst2}
\end{eqnarray}
Notice that in system (\ref{dynsyst1})-(\ref{dynsyst2}) the probabilistic nature of the model is captured by the \emph{process disturbance} $\bm{\omega}_t $ and \emph{measurement noise} $\bm{\nu}_t$, which are supposed to be sequences of
independent random variables with known probability density functions.

The UKF  does not directly approximate the nonlinear process and observation models, but exploits the nonlinear models, approximating the distribution of the state.
This is made possible by means of the \emph{scaled unscented transformation (SUT)} \cite{UKF}, which is a tool for computing the statistics of a random variable undergoing a nonlinear transformation.
Specifically, the state distribution is specified using a minimal set of deterministically chosen sample points.
Such sample points exactly provide the true mean and covariance of such a variable and, when propagated through the nonlinear transformation, 
they approximate the posterior mean and covariance accurately to the \emph{2nd order} for any nonlinearity. \textcolor{black}{For the reader's convenience, a brief review of the SUT is provided hereafter.}

 
 Let $\bm{x} \in \mathbb{R}^{n_x}$ be a random variable, with mean $\bm{\bar{x}}$ and covariance $P_{x}$, and $\bm{g}: \mathbb{R}^{n_x}\rightarrow \mathbb{R}^{n_y}$ an arbitrary nonlinear function.
The goal is to approximate the mean value $ \bm{\bar{y}}$ and covariance $P_y$ of  the variable  
$\bm{y}=\bm{g}(\bm{x})$.
  A set of $2n_x+1$ weighted samples or \emph{sigma points} $\mathcal{S}_i=\{\mathcal{W}_i, \bm{\mathcal{X}}_i\}_{i=0}^{2n_x}$ are chosen to completely represent the true mean and covariance of the variable $\bm{x}$, i.e.
   \begin{equation*}
    \begin{split}
     \bm{\mathcal{X}}_0\text{ }&= \bm{\bar{x}}\\
     \bm{\mathcal{X}}_i\text{ }&= \bm{\bar{x}}+(\sqrt{(n_x+k)P_x})_i \qquad i=1, \dots, n_x \\
     \bm{\mathcal{X}}_i\text{ }&= \bm{\bar{x}}-(\sqrt{(n_x+k)P_x})_i\qquad i=n_x+1, \dots, 2n_x \\
     \bm{\mathcal{W}}_0^{(m)}&= \lambda/(n_x+\lambda)\\
     \bm{\mathcal{W}}_0^{(c)}&= \lambda/(n_x+\lambda)+ (1-\alpha^2+\beta)\\
     \bm{\mathcal{W}}_i^{(m)}&=\bm{\mathcal{W}}_i^{(c)}= 1/[2(n_x+\lambda)] \qquad i=1, \dots, 2n_x,
     \label{sut3}
     \end{split}
   \end{equation*}
where: $ \lambda= \alpha^2(n_x+k)-n_x $; $\alpha>0$ provides one more degree of freedom to control the scaling of the
      sigma points and to avoid the possibility of getting a non-positive semi-definite covariance; $k\geq0$ is another scaling parameter; $\beta$ affects the weighting of the zero-\emph{th} sigma point.

      Each sigma point is then propagated through the function $\bm{g}(\cdotp)$ ($
        \bm{\mathcal{Y}}_i=\bm{g}(\bm{\mathcal{X}}_i)$, for $ i=0, \dots, 2n_x)$
        and the estimated mean and covariance of $\bm{y}$, as well as the cross-covariance between $\bm{x}$ and $\bm{y}$, are computed as follows:
        \begin{equation}
        \label{ut1}
        \begin{split}
        \bm{\bar{y}}&=\sum_{i=0}^{2n_x}{\mathcal{W}_i\bm{\mathcal{Y}}_i}\qquad
        P_{\bm{y}}=\sum_{i=0}^{2n_x}{\mathcal{W}_i(\bm{\mathcal{Y}}_i-\bm{\bar{y}})(\bm{\mathcal{Y}}_i-\bm{\bar{y}})^T}\\
         P_{\bm{x}\bm{y}}&=\sum_{i=0}^{2n_x}{\mathcal{W}_i(\bm{\mathcal{X}}_i-\bm{\bar{x}})(\bm{\mathcal{Y}}_i-\bm{\bar{y}})^T}.
        \end{split}
        \end{equation}
 
The Unscented Kalman Filter is  obtained by applying the SUT to the nonlinear functions $f_t$ and
$h_t$ in (\ref{dynsyst1})-(\ref{dynsyst2}). 

In practice, \textcolor{black}{in the UPF algorithm,} given the mean $\bm{x}_{t-1|t-1}^{(i)}$  and covariance $P_{t-1|t-1}^{(i)}$ at time $t-1$, as well as the mean and covariance of the process disturbance $\bm{\omega}_{t-1}$, application of the SUT
to the state transition equation (\ref{dynsyst1}) allows to compute an approximation of the predicted mean $\bm{x}_{t|t-1}^{(i)}$ and covariance $P_{t|t-1}^{(i)}$ at time $t$.
In turn, given  $\bm{x}_{t|t-1}^{(i)}$ and $P_{t|t-1}^{(i)}$ as well as the mean and covariance of the measurement noise $\bm{\nu}_t$,
application of the SUT to the measurement equation (\ref{dynsyst2}) allows  to provide an approximation of the predicted measurement mean $\bm{y}_{t|t-1}^{(i)}$ and covariance 
$S_{t}^{(i)}$
as well as of the state-measurement cross-covariance matrix $\Gamma_{t}^{(i)}$.  Then, the updated mean $\bar{\bm{x}}_t^{(i)}$ and covariance $P_{t|t}^{(i)}$ are obtained  by applying the standard Kalman filter correction step.


Since in practice it can happen that, after a few iterations, one of the normalized weights tends to 1, while the remaining weights tend to zero
(\textit{weight degeneration}), 
a selection or \emph{resampling} stage is usually included in the particle filtering algorithm, in order to eliminate samples with low importance weights and replicate samples with high importance weights.
Summing up, the resulting algorithm is reported in Table \ref{UPF}.
 \begin{table}
  \caption{The unscented particle filter}
  \label{UPF}
  \begin{algorithmic}
  \STATE
    \hrulefill\\
    \vspace{-4mm}
    \hrulefill\\
    \vspace{2mm}
  \STATE for $i=1, \dots, N$ draw the state particles $\bm{x}_{0|0}^{(i)}$ from the prior $p_{0|0}(\bm{x})$ and set 
                   $
                  P_{0|0}^{(i)}=\GiuliaAdd{P_{0}}$, and $\tilde{w}_{0}^{(i)}=1/N$\\
        \vspace{2mm}          
   \FOR{$t=1, 2,  \dots, $}
   \STATE{\bf 1) UKF prediction and correction}
\FOR{$i=1, \dots, N$}
  \STATE -{\bf Time update}: given $\{ \bm{x}_{t-1|t-1}^{(i)} , P_{t-1|t-1}^{(i)} \}$, compute $\{ \bm{x}_{t|t-1}^{(i)} , P_{t|t-1}^{(i)} \}$ by applying the SUT to the state transition equation (\ref{dynsyst1});
  \STATE -{\bf Measurement prediction}: given $\{ \bm{x}_{t|t-1}^{(i)} , P_{t|t-1}^{(i)} \}$, compute $\{ \bm{y}_{t|t-1}^{(i)} , S_{t}^{(i)} , \Gamma_{t}^{(i)} \}$ by applying the SUT to the measurement equation 
  (\ref{dynsyst2});
      \STATE -{\bf Measurement update}: set
      \begin{equation*}
         \begin{split}
           K_t^{(i)}= & \, \Gamma_t^{(i)} \left ( S^{(i)}_t \right )^{-1}\\                          \bm{\bar{x}}^{(i)}_t= & \, \bm{x}^{(i)}_{t|t-1}+K_t^{(i)} \,(\bm{y}_t-\bm{{y}}^{(i)}_{t|t-1})\text{ } \\
                                               P_{t|t}^{(i)}=& \, P^{(i)}_{t|t-1}-K_t^{(i)} S^{(i)}_t \left( K_t^{(i)} \right)^T\\              
                        \end{split}
                         \end{equation*} \ENDFOR
                         \STATE{\bf 2) Weight update}
                              \FOR{$i=1, \dots, N$}
                          \STATE sample from the proposal distribution:
                             \begin{equation*}
                             \hat{\bm{x}}_t^{(i)}\sim q^{(i)} (\cdot | \bm{y}^t) = 
                             \mathcal{N} (\cdot;\bar{\bm{x}}_t^{(i)}, P_{t|t}^{(i)})
                             \end{equation*}               

                                           \STATE evaluate and normalize the importance weights:
                                           \begin{equation*}
                                             w_t^{(i)} = \tilde w_{t-1}^{(i)} \frac{\ell_t(\bm{y}_t|\bm{\hat{x}}_{t}^{(i)})\text{ }\varphi_{t|t-1}(\bm{\hat{x}}_{t}^{(i)} |\bm{x}_{t|t-1}^{(i)})}{ \mathcal{N} ( \hat{\bm{x}}_t^{(i)};\bar{\bm{x}}_t^{(i)}, P_{t|t}^{(i)})}
                                            \end{equation*}
                                            \begin{equation*}
                                            \tilde{w}_t^{(i)} = w_t^{(i)} / \sum_{j=1}^N   w_t^{(j)}  
                                           \end{equation*}
                               \ENDFOR
                         \STATE{\bf 3) Resampling}
                                      \FOR{$i=1, \dots, N$}
                                      
                                      \STATE draw $j\in\{1, \dots, N\}$ with probability $\tilde{w}_{t}^{(j)}$
and set: \textcolor{black}{ 
     \begin{equation*}	\bm{x}^{(i)}_{t|t}=\bm{\hat {x}}_{t}^{(j)}\quad P^{(i)}_{t|t} = P^{(j)}_{t|t} \quad
                                      	\tilde{w}_{t}^{(i)}=\frac{1}{N}
                                      	\end{equation*} }
                                  
                                      \ENDFOR


    \ENDFOR
    \\\hrulefill\\
    \vspace{-4mm}
    \hrulefill
  \end{algorithmic}
  \end{table}

\section{Problem formulation}\label{prob}


\textcolor{black}{The object to be localized is assumed to be static during the measurement collection. This \LorenzoAdd{assumption is common to other works \cite{markov, Petr, Petr2,actionsel}} and is realistic, for instance, if the object is very heavy or is stuck 
on a support preventing any possible movement.
Hence,} the goal of the $6$-DOF object tactile localization problem is to estimate in real-time the pose  \textcolor{black}{$\bm{x} \in \mathbb{R}^6$} of an object $\mathcal{O}$ of known shape, 
on the basis of \textcolor{black}{the tactile measurements $\bm{y}^t=\{\bm{y}_1, \dots, \bm{y}_t\}$ collected up to the current time instant $t$}. The minimal pose representation  of the object is given by the 6-dimensional state vector \textcolor{black}{ $\bm{x}$}, consisting of the coordinates of the center of the reference system attached to the object and the three Euler angles representing the orientation, i.e.
 \textcolor{black}{ \begin{equation}
  \bm{x}=
  \begin{bmatrix}
 x,& y,& z,& \phi,& \theta, &\psi
  \end{bmatrix}^T.
  \end{equation}}
  The measurements are collected by touching the object with the end effector of the robot. Each measurement $\bm{y}_{t}$ consists of the acquired Cartesian position of the contact point, i.e.:
  \begin{equation}
  \bm{y}_{t}=
  \begin{bmatrix}
  x_{t,p},& y_{t,p}, & z_{t,p}
  \end{bmatrix}^T.
  \end{equation}
\indent It is worth noticing that the exploited measurements consist only of tridimensional contact point vectors. 
\textcolor{black}{Notice also that, while for ease of presentation it is assumed that a single measurement consists of a single contact point, the proposed approach
is well-suited to being extended to consider measurements consisting of multiple contact points (corresponding to different fingertips touching the object).
This would simply amount to processing, at each time $t$, a measurement vector of size $3 n_t$, $n_t$ being the number of fingertips touching the object at that time.}
Finally, notice that, in the sequel, all measurements and the object pose will be assumed to be expressed in the same, fixed, reference system. 

\subsection{\textcolor{black}{Considerations on the motion model}}

\textcolor{black}{Since the object is assumed to be static, the $6$-DOF object tactile localization problem is basically a static parameter estimation problem. 
In this respect, it is well known that the use of particle filtering techniques for estimating static parameters requires special care, because a direct application of these techniques
to the constant state equation $\bm{x}_{t+1} = \bm{x}_t$, corresponding to the Markov transition density $\varphi_{t+1|t} (\bm{x} | \bm{\xi}) = \delta(\bm{x}-\bm{\xi})$, would incur in the so-called weight-degeneracy phenomenon.
Many solutions have been proposed in the literature to circumvent such a problem, see for instance \cite{Liu2001} and the references therein.
A simple but effective approach consists in adding an artificial dynamic noise on the static parameter by considering a state-transition equation of the form
\begin{equation}\label{eq:motion}
\bm{x}_{t+1}=\bm{x}_{t}+\bm{\omega}_{t},
\end{equation}
where $\bm{\omega}_{t}$ is the artificial dynamic noise which is modeled as a Gaussian random variable with zero mean and covariance $Q_t$. 
The idea is that the artificial evolution provides a mechanism for generating at each time instant new particles with a sufficiently diffuse distribution. 
In this paper, a time-invariant covariance matrix is used, i.e. $Q_t =Q$, since it proves effective in the considered case studies. However
more elaborated solutions can be easily incorporated within the proposed algorithm \cite{Liu2001}. \\ \indent Some considerations on the possibility of extending the approach to the case of moving object localization are provided in Remark \ref{rem:TV}.
}

%

\subsection{Measurement Model}

In order to apply the UPF to the tactile localization problem under investigation, it is necessary to define the measurement model both in terms of a likelihood function $\ell_t(\bm{y}_t|\bm{x}_t)$ and of a measurement function $\bm{h}_t(\cdotp,\cdotp)$.  The proposed \emph{likelihood function} is based on the so-called \emph{proximity model}, in which the measurements are considered independent of each other and corrupted by Gaussian noise.  For each measurement, the likelihood function depends on the distance between the measurement and the object, hence the name \textquotedblleft proximity". This model is the adaptation of the likelihood proposed in \cite{Petr} to the case of contact point measurements only.

Let the 3D object model be represented by a polygonal mesh consisting of faces \{$f_i$\}. For each face $f_i$, let $\ell_{t,i}(\bm{y}_t|\bm{x}_t)$ be the 
 likelihood of the measurement $\bm{y}_t$ relative to that face when the object is in the pose $\bm{x}_t$.
Then, the likelihood of the measurement is  defined as the maximum likelihood over all faces, i.e.
  \begin{equation}
  \label{likelihood}
  \ell_t(\bm{y}_t|\bm{x}_t) \propto \max_i{\ell_{t,i}(\bm{y}_t|\bm{x}_t)},
 \end{equation}
apart from  a normalizing factor which, however, is independent of the state $\bm{x}_t$ and needs not necessarily be computed. 
     
  Each likelihood is assumed to be Gaussian, with variance $\sigma_p^2$,  and can be computed as follows:
     \begin{equation}
     \ell_{t,i}(\bm{y}_t|\bm{x}_t)=\frac{1}{\sqrt{2 \pi}\sigma_{p}}
		\exp \left(-\frac{1}{2} \,\, \frac{d_i(\bm{y}_{t},\bm{x}_t)^2}{\sigma_{p}^2} \right),
     \end{equation}
     where the quantity $d_i(\bm{y}_{t},\bm{x}_t)$ is the shortest Euclidean distance of $\bm{y}_{t}$ from the face $f_i$ when the object is in the pose $\bm{x}_t$. For instance, supposing that $f_i$ is the representation of the
$i$-th face in the object reference system, the distance $d_i(\bm{y}_{t},\bm{x}_t)$ can be computed as
\[
d_i(\bm{y}_{t},\bm{x}_t) = \min_{\bm{p} \in f_i} \| \bm{y}_{t}^{\bm{x}_t} - \bm{p} \|,
\]
where $\| \cdot \|$ is the Euclidean norm and $\bm{y}_{t}^{\bm{x}_t}$ denotes the transformation of the measurement $\bm{y}_{t}$ using the roto-translation matrix corresponding to the state $\bm{x}_t$.

Notice that the considered measurement model does not take \emph{negative information} into account. 
In other words, the points of the  search space exploited to compute the likelihood function are only the ones  on the object surface  touched during the collection of  measurements, while
the information provided by the lack of contact in some sub-regions of the search space is not taken into account in the likelihood function. 
Even if the negative information can also support object localization, it
is not exploited in this method  in order to keep the computational complexity moderate.
   
  As previously pointed out, the use of the UPF requires also the definition of a \emph{measurement function}, namely a mathematical mapping giving the measurement $\bm{y}_t$ as a function of  the current state $\bm{x}_t$ and a measurement noise $\bm{\nu}_t$, see (\ref{dynsyst2}). For the sake of simplicity, a measurement equation with additive noise is taken into account, i.e.,
   \begin{equation}\label{eq:additive}
   \bm{y}_t =\bm{h}_t(\bm{x}_t) + \bm{\nu}_t \, .
   \end{equation}
   In particular, the measurement function is required to compute the Scaled Unscented Tranform (SUT) in the measurement prediction step of the Unscented Kalman Filter \GiuliaDelete{(Algorithm \ref{UPF})}.

It is important to highlight how the definition of a measurement equation is different from the one of  a likelihood function: given the state and the measurement noise, the measurement equation provides a measurement value - a contact point in the present case - whereas the likelihood function is proportional to the probability of having a certain measurement
for a given state. 

Tactile sensors are atypical sensors from this standpoint. 
In fact, typical sensors, e.g. radars, are characterized by a mathematical relationship between the current state of the object and the provided measurement: given the state of the object, the measurement of the object position and orientation supplied by the sensor remains unchanged (neglecting the measurement noise).

On the other hand, the employment of tactile sensors makes the scenario quite different. The measurement is given by the tactile sensor pose itself, i.e., the forward kinematics of the end effector of the robot,  only if the robot actually touches the object. Thus, if the object is in a generic state and the sensor in a specific pose, it cannot be taken for granted that such a configuration provides a contact measurement. Moreover,  the sensor  moves during the measurement collection, while the object is motionless.  It is not possibile to predict unambiguously the measurement value without a model of the sensor motion: given the pose of the sensor  and the object  distance from it, the predicted measurement is not unique, since  the sensor could  touch the object in different points.

Nevertheless, in order to compute a predicted measurement for each possible configuration $\bm{x}$, it is necessary to define a measurement equation capable of handling 
also the case in which there is no actual contact between the sensor and the object in the considered pose $\bm{x}$ (in particular, the sigma point of the $i$-th predicted particle). Further, the predicted measurement should be consistent with the proximity-based likelihood (\ref{likelihood}).

To this end, it is useful to provide an alternative interpretation of the likelihood (\ref{likelihood}). 
Notice first that, due to the measurement noise, the measurement $\bm{y}_t$ does not represent the actual contact point between
the sensor and the object, which however will be in the neighborhood of $\bm{y}_t$. 
The proximity model assumes that the actual contact point is the point on the object surface which is closest to the
measurement $\bm{y}_t$. \textcolor{black}{In fact, equation (\ref{likelihood}) can be rewritten as
\begin{equation}\label{eq:likelihood}
 \ell_t(\bm{y}_t|\bm{x}_t) \propto  \exp \left(-\frac{1}{2 \sigma_p^2} \,\, \left\| \bm{y}_t -  \bm{h}_t(\bm{x}_t) \right\|^2  \right)
\end{equation}
where 
\begin{equation}
\bm{h}_t \left( \bm{x}_t \right) = \arg \displaystyle{\min_{\bm{p} \in \partial \mathcal{O}^{\bm{x}_t}}} \left\| \bm{y}_t - \bm{p} \right\|
\label{eq:meas}
\end{equation}
and 
$\partial \mathcal O^{\bm{x}_t}$ represents the object boundary in the pose $\bm{x}_t$ with respect to the robot reference system.}
Then, the likelihood of the measurement $\bm{y}_t$ depends on its distance from such a hypothetical contact point according to a Gaussian distribution.
Accordingly, given a configuration $\bm{x}_t$, the corresponding predicted measurement is selected as the point of the object surface  which is closest to the measurement $\bm{y}_t$. 
 \textcolor{black}{Such a choice turns out to be consistent with the proximity likelihood model. In fact, by taking the additive measurement noise $\bm{\nu}_t $ in (\ref{eq:additive}) as a Gaussian random variable with zero-mean and covariance $\sigma_p^2 I$, with $I$ the identity matrix,
 it is an easy matter to see that (\ref{eq:additive}) and (\ref{eq:meas}) give rise precisely to a likelihood of the form (\ref{eq:likelihood}). }


\section{The memory unscented particle filter}\label{mupfsec}

The main challenges of the $6$-DOF tactile localization problem are its dimension ($6$-DOFs), its  multimodal nature, and the fact that individual measurements are relatively uninformative, since they are tridimensional vectors in a $6$-DOF space.
In particular, the latter fact implies that the standard UPF algorithm is not well suited to this problem.
In fact, Algorithm \ref{UPF} uses, at each time instant $t$, only the current measurement $\bm{y}_t$ in order to compute the importance weights $w_t^{(i)}$. 
\textcolor{black}{However, since a single contact point measurement is unable to completely characterize the object's pose 
(lack of observability), the standard weights} \LorenzoDelete{ need not} \LorenzoAdd{do not}
 provide enough information to understand which particles must be replicated and which ones must be eliminated in the subsequent resampling step.
Thus, performing the standard resampling  step - and then discarding some particles - on the basis of such weights is  problematic: some potential representative particles could be cut off and the algorithm could limit the search to wrong sub-regions. 

In order to overcome such a drawback, this paper proposes a novel variant of the UPF, referred to as Memory UPF (MUPF). 
\textcolor{black}{The idea  is to use also past measurements to update particle weights so as to preserve their ability to characterize particle goodness. Since the object is static, all the measurements refer to the same pose and, in principle, 
at each time $t$ all the measurements $\bm{y}^t$ collected up to the current time could be used to compute the importance weights.
However, this solution would entail a computational effort growing in time. To avoid such a growth of complexity, the
proposed approach follows a {\em moving window} strategy, i.e., by using, at each time instant, a sliding window
consisting of the most recent $m$ measurements. In this way, at each time instant, the weight computation
requires $O(N m)$ likelihood evaluations, and the size $m$ of the sliding window can be chosen according to the available
computational capabilities.
}

\textcolor{black}{
In practice, the 
particles $\{\hat{\bm{x}}^{(i)}_t\}_{i=1}^N$ and the set of  independent measurements $\{\bm{y}_1, \dots, \bm{y}_t\} $,  collected up to the current instant $t$, are used to compute the  weights by:
   \begin{equation}
   \label{weightt}
   w_{t}^{(i)}=
   \frac{\tilde{w}_{t-1}^{(i)}\cdotp \prod_{k=\bar{k}(t)}^t\text{ } \ell(\bm{y}_k|\hat{\bm{x}}_t^{(i)})}
   { \mathcal N(\hat{\bm{x}}_t^{(i)};  \bar{\bm{x}}_t^{(i)}, P^{(i)}_{t|t})}
	, 
   \end{equation} 
 \begin{equation}\label{weightt:norm}
 \tilde{w}_t^{(i)} = w_t^{(i)} / \sum_{j=1}^N   w_t^{(j)}
 \end{equation} 
  for $ i=1, \dots, N$, where
         \begin{equation}
         \bar{k} (t)=
         \begin{sistema}
				 \begin{array}{ll}
         t-m+1, & \text{if } t-m+1 \ge 1\\
         1, & \text{otherwise}.
				\end{array}
    \end{sistema}
         \end{equation}
}

\textcolor{black}{Of course, the reuse of measurements in the update of the particles' weights modifies the nature of the approximation, and hence special care needs to be taken in order to retrieve the pose estimate in a theoretically sound way. To see this, observe preliminarily, that
the addressed problem is  inherently of a multimodal nature, since in the presence of symmetries in the object, 
there might exist multiple values of $\bm{x}$ compatible with the measurements. 
Then, taking the expected value as estimate is not meaningful. Instead, a maximum \emph{a posteriori} probability (MAP) criterion can be followed by taking as pose estimate at time $t$ 
the corrected particle $\bm{\hat{x}}_t^{(i)}$ corresponding to the
highest value of the  estimated posterior distribution \cite{particlebased} .
}

\textcolor{black}{Recalling that each corrected particle can be considered corresponding to  a Gaussian distribution with mean 
$\bm{\bar{x}}_t^{(i)}$ and covariance $P_{t|t}^{(i)}$,  
one might be tempted to take as estimated posterior $\hat p_{t|t}(\cdotp)$ the function
\begin{equation}\label{eq:est:post}
 \hat p_{t|t}(\bm{x}) = \sum_{i=1}^{N}{\tilde w_t^{(i)}} \, \mathcal{N}(\bm{x};\bm{\hat{x}}_t^{(i)},P^{(i)}_{t|t}).
\end{equation}
Unfortunately, such a choice would not be theoretically sound due to the multiple use of measurements in the weight computation.  In this respect, notice first that, since the object is static, the $6$-DOF localization problem is a parameter estimation problem and, hence, the true posterior $p_{t|t}(\cdotp)$ at time $t$ takes the form
\begin{equation}
p_{t|t}(\bm{x})  \propto  \prod_{k=1}^t\text{ } \ell(\bm{y}_k|\bm{x})  \, p_0(\bm{x}),
\end{equation}
}
where $p_0(\cdot)$ is a PDF reflecting the prior knowledge on the object configuration.
Since at each time instant multiple measurements are used in the weight computation, the
estimated posterior $\hat p_{t|t}(\cdotp)$ does not approximate the true one $p_{t|t}(\cdotp)$ but instead \LorenzoAdd{it approximates} the PDF
\textcolor{black}{
\begin{equation}
\tilde p_{t|t}(\bm{x})\propto \prod_{k=1}^{\bar k (t)-1} \ell(\bm{y}_k|\bm{x})^{m} \,
\prod_{k=\bar{k}(t)}^t\text{ } \ell(\bm{y}_k|\bm{x})^{t+1-k} \, p_{0|0}(\bm{x}),
\end{equation}
}
where $p_{0|0}(\cdot)$ is the prior density used in the generation of the initial particles,
\textcolor{black}{thus introducing an undesired warp in the form of the estimated posterior PDF.}

\textcolor{black}{This drawback can be circumvented by computing special weights $\bar w^{(i)}_t$, used only
for the purpose of pose estimation extraction but not propagated in the recursion. 
In fact, by setting
   \begin{equation}
   \label{weightt2}
   w_{t}^{(i)}=
   \frac{\tilde{w}_{t}^{(i)}\cdotp \prod_{k=\bar{k}(t)}^t\text{ } \ell(\bm{y}_k|\hat{\bm{x}}_t^{(i)})^{m-t+k-1}}
   { \mathcal N(\hat{\bm{x}}_t^{(i)};  \bar{\bm{x}}_t^{(i)}, P^{(i)}_{t|t})}
	, 
   \end{equation} 
 \begin{equation}\label{weightt2:norm}
 \bar{w}_t^{(i)} = w_t^{(i)} / \sum_{j=1}^N   w_t^{(j)}
 \end{equation} 
for $ i=1, \dots, N$, and using the estimated posterior 
\begin{equation}\label{eq:est:post2}
\hat p_{t|t} ( \bm{x}) = \sum_{i=1}^{N}{\bar w_t^{(i)}} \, \mathcal{N}(\bm{x};\bm{\hat{x}}_t^{(i)},P^{(i)}_{t|t})
\end{equation}
in place of (\ref{eq:est:post}), it turns out that such a 
$\hat p_{t|t}(\cdotp)$
}
approximates the PDF
\begin{equation}\label{eq:PDF:m}
\bar p_{t|t}(\bm{x}) \propto \prod_{k=1}^t \text{ }\ell(\bm{y}_k|\bm{x})^{m} \, p_{0|0}(\bm{x}) 
\end{equation}
so that all measurements provide the same contribution to the estimation problem.
Then, by choosing $p_{0|0}(\bm{x}) \propto p(\bm{x}_0)^m $, we obtain $\bar p_{t|t}(\bm{x}) \propto  p_{t|t}^m(\bm{x})$ which implies that $\bar p_{t|t}(\bm{x})$ and  $p_{t|t}(\bm{x})$ share the same maximum points.
In turn, this implies that application of the MAP estimation criterion to $\bar p_{t|t}(\bm{x})$ is equivalent to computing the MAP estimate according to $p_{t|t}(\bm{x})$.
\textcolor{black}{These considerations allow concluding that, with the choice $p_{0|0}(\bm{x}) \propto p_0(\bm{x})^m $, the MAP estimate $\bm{\hat{x}}_t$
corresponding to the particle with the maximum a posteriori probability  according to (\ref{eq:est:post2})
   \begin{eqnarray}\label{eq:pose}
  &\bm{\hat{x}}_t=\arg \displaystyle{\max_{j}} \,\,  {\hat p_{t|t}(\bm{\hat{x}}_t^{(j)})}=\\  
  &=\arg \displaystyle{\max_{j}} \sum_{i=1}^{N}{\bar w_t^{(i)}}\mathcal{N}(\hat{\bm{x}}_t^{(j)};\bm{\hat{x}}_t^{(i)},P^{(i)}_{t|t}).
   \end{eqnarray}
is coherent with the true posterior PDF.}

\textcolor{black}{ 
\begin{remark} The fact that (\ref{eq:est:post2}) approximates (\ref{eq:PDF:m}) can be shown by noting that $\bar p_{t|t}(\bm{x}) $ can be decomposed as follows
\begin{eqnarray*}
\lefteqn{\bar p_{t|t}(\bm{x})  \propto   \prod_{k=\bar{k}(t)}^t \text{ }\ell(\bm{y}_k|\bm{x})^{m-t+k-1} } \\
&& {} \times    
\prod_{k=\bar{k}(t)}^t\text{ } \ell(\bm{y}_k|\bm{x})^{t+1-k} \, \prod_{k=1}^{\bar k (t)-1} \ell(\bm{y}_k|\bm{x})^{m} 
\, p_{0|0}(\bm{x}) \\
&& {}  = \prod_{k=\bar{k}(t)}^t \text{ }\ell(\bm{y}_k|\bm{x})^{m-t+k-1} \, \tilde p_{t|t} (\bm x)
\end{eqnarray*}
which precisely corresponds to the weight update in (\ref{weightt2}).
\end{remark}
}
 \begin{table}
 	\caption{The memory unscented particle filter}
 	\label{mupf}
 	\textcolor{black}{
 		\begin{algorithmic}
 			\STATE
 			\hrulefill\\
 			\vspace{-4mm}
 			\hrulefill\\
 			\vspace{2mm}
 			\STATE for $i=1, \dots, N$ draw the state particles \GiuliaAdd{$\bm{x}_{0|0}^{(i)}$} from the prior $p_{0|0}(\bm{x})$ and set $
 			P_{0|0}^{(i)}=\GiuliaAdd{P_{0}}$ and $\tilde{w}_{0|0}^{(i)}=1/N$\\
 			\vspace{2mm}
 			\FOR{$t=1, 2, \ldots$} 
 			\STATE{\bf 1) UKF prediction and correction}
 			\FOR{$i=1, \dots, N$}
 			\STATE -{\bf Time update}: set $\bm{x}_{t|t-1}^{(i)} = \bm{x}_{t-1|t-1}^{(i)}$ and $P_{t|t-1}^{(i)} = P_{t-1|t-1}^{(i)} + Q$;
 			\STATE -{\bf Measurement prediction}: like in Table I;
 			\STATE -{\bf Measurement update}: like in Table I; 		\ENDFOR
 			\STATE{\bf 2) Weight update}
 			\FOR{$i=1, \dots, N$}
 			\STATE sample from the proposal distribution:
 			\begin{equation*}
 			\hat{\bm{x}}_t^{(i)}\sim q^{(i)} (\cdot | \bm{y}^t) = 
 			\mathcal{N} (\cdot;\bar{\bm{x}}_t^{(i)}, P_{t|t}^{(i)})
 			\end{equation*}               
 			\STATE evaluate and normalize the modified importance weights via (\ref{weightt}) and (\ref{weightt:norm});
 			\STATE 
 			\ENDFOR\\
 			\STATE{\bf 3) Estimated pose extraction (optional)}
 			\FOR{$i=1, \dots, N$}
 			\STATE evaluate and normalize the importance weights via (\ref{weightt2}) and (\ref{weightt2:norm});
 			\ENDFOR\\
 			\STATE compute the estimated pose $\bm{\hat{x}}_t$ via (\ref{eq:pose});
 			\STATE{\bf 4) Resampling}
 			\FOR{$i=1, \dots, N$}
 			\IF{$t>t_0$} 
 			\STATE 
 			\STATE draw $j\in\{1, \dots, N\}$ with probability $\tilde{w}_{t}^{(j)}$
 			\STATE then set: \begin{equation*}
 			\bm{x}^{(i)}_{t|t}=\bm{\hat{x}}_{t}^{(j)}
 			\qquad P_{t|t}^{(i)} = P_{t|t}^{(j)} 
 			\qquad \tilde{w}_{t}^{(i)}=\frac{1}{N}   	
 			\end{equation*} 	                                                                          
 			\ELSE 
 			\STATE set :
 			\begin{equation*}
 			\bm{x}^{(i)}_{t|t}=\bm{\hat{x}}_{t}^{(i)} \qquad  \tilde{w}_{t}^{(i)}=\frac{1}{N}
 			\end{equation*} 	                                 	                             	                                  	                                  
 			\ENDIF
 			\ENDFOR		
 			\ENDFOR
 			\\\hrulefill\\
 			\vspace{-4mm}
 			\hrulefill
 		\end{algorithmic}
 	}
 \end{table}
\textcolor{black}{A further modification, as compared to the standard UPF, pertains to the resampling step}. 
Since in the first iterations only few measurements are available (thus providing insufficient information), all the particles are retained so as to  account for more possibile solutions, in accordance with the multimodal nature of the problem. This amounts to skipping the standard resampling step for a certain number \textcolor{black}{$t_0$} of initial time instants (in the experimental results reported in the following sections, for the first two time instants). 
The degeneration of the weights in the first iterations is avoided by setting the weights of all particles equal to $1/N$.

\textcolor{black}{ Summing up, the proposed MUPF algorithm is shown in Table \GiuliaAdd{\ref{mupf}}.}
The term \emph{Memory}, in the name of the proposed algorithm, is due to the computation of the weights: at each iteration a non-decreasing number of measurements is exploited to evaluate the likelihood function.
\textcolor{black}{Notice also that the computation of the weights $\bar w^{(i)}_t$ is optional (since they are not used
in the time propagation from $t$ to $t+1$) and can be limited only to those time instants in which
one wants to extract an estimate $\bm{\hat{x}}_t$ of the object's pose from the approximated posterior.
}

\textcolor{black}{ 
\begin{remark}\label{rem:TV}
While the considered framework deals with static objects, the proposed algorithm is well-suited to being extended
to the case of moving objects since it is based on Bayesian filtering and is inherently recursive in nature.
When the object is not static, however, the use of a sliding window of the most recent measurements in the weight computation
requires some caution because the past measurements do not refer to the current pose. In principle, 
this problem can be circumvented by considering particle states consisting of the whole object trajectory in the sliding window  (similarly to what happens in particle-filtering-based solutions to the SLAM problem) so that the likelihood, with respect
to the measurements in the sliding window, can be correctly computed. Further, when the object is static, a simple motion
model like (\ref{eq:motion}) makes sense only to model small random movements caused by probing. For truly moving objects 
(for example a rolling ball), more complex motion models are required including also the object velocity.
Of course, the main challenge in this case is the increased complexity due to such modifications.
Such generalizations are left for future research. 
\end{remark}
}


 \section{Algorithm validation with simulated measurements}\label{sim}
 In order to evaluate the performance of the developed Memory Unscented Particle Filter (MUPF), a C++ implementation of MUPF has been tested via simulations on different objects and collections of measurements. 
The tests have been run on a Linux platform, with a quadcore $3.40 \, GHz$ processor.
The developed code, the exploited measurements and the reconstructed object models can be downloaded from  \textit{github}\footnote{ DOI:10.5281/zenodo.163860.}.
 
 \subsection{Simulation setup}\label{simsetup}
 \indent The  simulation setup consists of five objects: a rectangular box, a
 tetrahedron,  a cleaner spray, a robot toy and a safety helmet (Fig. \ref{simobjects}).\\
 \indent The mesh models of the first two objects, having a simple geometrical shape, are built from ruler measurements whereas the other three more complicated objects are approximated by triangular mesh models, reconstructed via image processing algorithms. In particular, the  mesh models  of the cleaner spray and safety helmet\GiuliaDelete{, each consisting of 250 triangular faces,} are obtained from  360 degree point clouds reconstructed with the RTM toolbox\footnote{Recognition Tracking and Modelling of Objects,  by ACIN of Technische Universit\"{a}t Wien, http://www.acin.tuwien.ac.at/forschung/v4r/software-tools/rtm/.} \cite{rtm}. The RTM toolbox merges together several partial 3D models - i.e. different views of the object - captured by rotating the object in front of a RGB-D camera    and, in a few seconds, provides a 360 degree point cloud of the object. Conversely, the more complex  point cloud of the robot toy is retrieved by making use of the AutoDesk 123d catch application\footnote{http://www.123dapp.com/catch.} that, in several tens of minutes, processes different   object photos taken from different views with a smartphone.  Thus,  the triangular mesh models of the three objects are extracted by applying the Poisson Surface Reconstruction algorithm \cite{poiss} to the merged point clouds. \GiuliaDelete{The robot toy mesh model  consists of 750 triangular faces. }The complete pipeline for model reconstruction is outlined in Fig. \ref{obj4}.\\
 \begin{figure}
 	\centering{\includegraphics[scale=0.35]{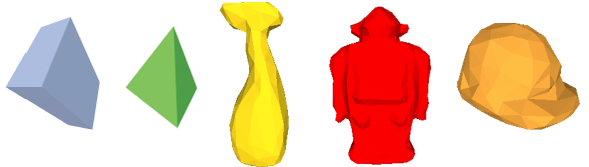}
 		\caption{Simulation setup objects. From left to right: a rectangular box ($0.1 \times 0.3 \times 0.2 $ [m]), 
 			a tetrahedron (equilateral triangular basis with the side of 0.33 [m] $\times$ height of 0.2 [m]), a  cleaner spray (approximately $0.23 \times 0.08 \times 0.05 $ [m]), a robot toy ($0.23 \times 0.09 \times 0.06 $ [m]), and  a safety helmet (nearly a half-sphere with radius $0.1$ [m]) \label{simobjects}}.		
 	}
 \end{figure}
 \begin{figure}
 	\centering
 	\includegraphics[scale=0.25]{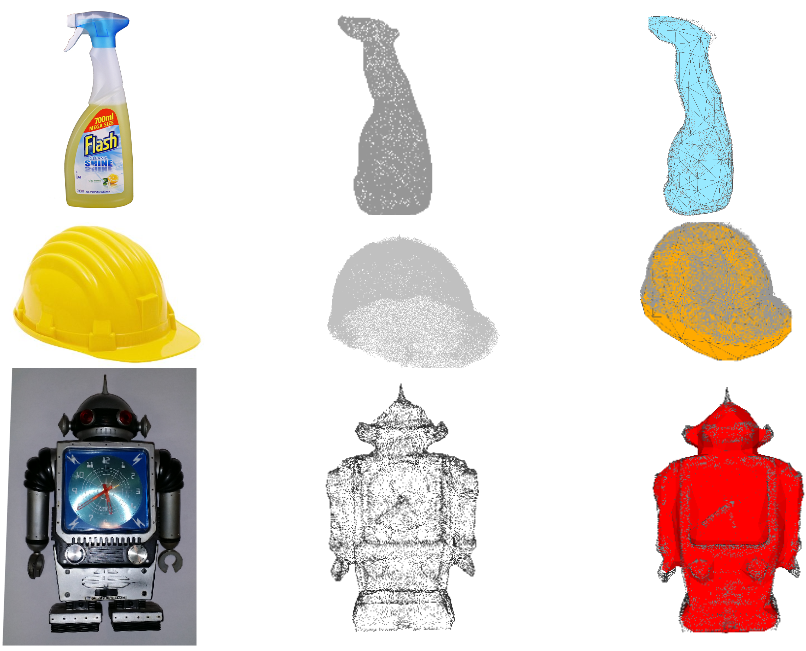}
 	\caption{Pipeline for  real object modelling. From left to right: real objects, 360 degree point clouds (obtained with a RGB-D camera and the RTM toolbox for the cleaner spray and the safety helmet, and with 40 photos from different views and the Autodesk 123d catch app for the robot toy), triangular mesh models matching the point clouds, computed by using the Poisson surface reconstruction. On the top: the cleaner spray,  whose mesh model consists of 250 faces. In the middle: the safety helmet, featured by a  mesh model of 250 faces. On the bottom: the robot toy, whose mesh model is made up of 750 faces. \label{obj4}}
 \end{figure}
 \indent The  contact point measurements exploited in the simulation tests are drawn by non-uniformly sampling random points on a subset  of  3D model faces.\\
 \indent  The MUPF algorithm requires setting the following parameters: 
\textcolor{black}{the artificial process noise covariance matrix $Q$}; 
the measurement noise covariance \GiuliaAdd{$\sigma_p$} characterizing sensor accuracy; 
the initial covariance matrix $P_0$ to quantify the initial uncertainty and, hence, the extent of the search region;
the parameters of the unscented transformation $\alpha, \beta,k$;
the number of particles $N$;
the length $m$ of the measurement window for the importance weight update. \\
 \indent \GiuliaAdd{As preliminary tests, t}he parameters are kept constant, as shown in Table \ref{MUPFpar}. 
In particular, the chosen matrix $Q$ \textcolor{black}{is such that the artificial process disturbance spreads the particles}
with standard deviations of $1 \, cm$ in position and about $5$ degrees in rotation. 
Conversely, the covariance \GiuliaAdd{$\sigma_p$} assumes that the measurements of the end-effector position are affected by an error with standard deviation of $1 \, cm$ in all Cartesian coordinates. Finally, the initial matrix $P_0$ indicates an initial uncertainty of $0.4 \, m$ in position and $360$ degrees in orientation. \GiuliaAdd{The initial particles $\bm{x}_{0|0}^{(i)}$ for $i=1, \dots, N$ are drawn from the prior distribution $\mathcal{N}(\bm{x}_0|P_0)$, where $\bm{x}_0$ is arbitrarily chosen (a 6D null vector in our tests)}.  The choices of Table \ref{MUPFpar} have proven effective in all the considered simulations, thus indicating that the proposed algorithm
works over a broad range of problems without a case-by-case parameter tuning.
\\
 \indent It is worth pointing  out how the exploitation of the UKF step in the UPF allows to considerably reduce the number of particles to $N=700$ (with a standard particle filter it would be in the order of $N=10^6$ for a $6$-DOF problem). \GiuliaAdd{Section \ref{analysis} provides a detailed analysis about the parameters influence on MUPF performance.}
 \begin{table}[h!]
 {\centering
 \caption{Parameter set for the MUPF\label{MUPFpar}}}
  \begin{tabular*}{\columnwidth}{c c} 
  \hline
  \hline
   &\\
 $Q$& diag([10$^{-5}$, 10$^{-5}$, 10$^{-5}$, 10$^{-4}$, 10$^{-4}$, 10$^{-4}$]) [m], [rad]\\
  $P_0$&  diag([0.04, 0.04, 0.04,  $\pi^2$, $(\pi/2)^2$, $\pi^2$]) [m], [rad]\\
   $\sigma_p$&  $10^{-4}$[m]\\
   $\alpha$ & 1  \\
     $k$ &  2  \\
      $\beta$ &30  \\
       $N$ & 700  \\
       &\\
         \hline
         \hline
    \end{tabular*}
    \end{table}\\
 \subsection{Performance evaluation}
 The performance of the proposed algorithm is assessed in terms of both effectiveness and execution time, since the ultimate aim of this work is a real-time application of the algorithm. \\
\indent In this respect,  algorithm \emph{reliability} is measured in terms of number of successes among trials\GiuliaDelete{ in both simulation and experimental tests}, where 
a trial is considered failed whenever the estimated pose is substantially different from the real one. \\
\indent In  simulation tests, successes and failures can be discriminated by computing the distance between the estimated and the true object poses, since  the knowledge of the latter is available. The situation is different in real experiments, wherein the true pose is often difficult (if not impossible) to be measured. In this case, the distinction between a successful or a failed trial is necessarily accomplished by the user by visually inspecting that the solution  found by the algorithm is consistent with the real pose of the object.
 In the successful cases, a numerical evaluation of the localization can be done by relying merely on measurements without the need of the ground truth. This choice is by far preferable (sometimes the only viable solution) for an experimental assessment.\\
\indent These considerations suggest the definition of the following \emph{performance index}:
\begin{equation}
\mathcal{I}_L=\frac{1}{L}\sum_i^L{d_i},
\label{pi}
\end{equation}
where \GiuliaAdd{$L$} is the total number of collected measurements and $d_i$ the distance between the $i$-th measurement and the object in the estimated pose. 
In other words, given the set of measurements and the estimated pose, the proposed performance index is the average of the distances between each measurement and the object in the estimated pose. \\
\indent The performance index  $\mathcal{I}_L$ has  been adopted to evaluate the localization quality in 
simulation \GiuliaAdd{(together with the standard localization error measured as distance of the final estimated pose
from the ground truth)
and experimental tests, for the reasons listed below. }
First, the index $\mathcal{I}_L$ is the only viable solution for the experimental tests, wherein the real pose cannot typically be known or measured with sufficient accuracy. 
\textcolor{black}{Secondly, the use of a common error index for both simulation and experimental tests, makes easier the comparison between the two cases. 
Third, if simulation tests are carried out with noiseless measurements and a sufficient number of informative measurements is collected, 
then the performance index $\mathcal{I}_L$ can be related to the distance between the estimated and the true object poses, in the sense that 
$\mathcal{I}_L$ vanishes for large $L$ if and only if the two poses coincide.
Finally, the index $\mathcal{I}_t$ is easily computable on-line at each time $t$ and could therefore be monitored in order to understand 
when to stop localization of the current object.}
As a further benefit, (\ref{pi}) provides a synthetic (scalar)  indicator of the pose error, in terms of linear displacement (measured in units of length). 
Thus, the index computation is not affected by the problems related to the computation of angular displacements. \\
\indent Nevertheless, it is worth  pointing out that when the measurements are too inaccurate, the index (\ref{pi}) can be non-informative and the evaluation of the algorithm performance would necessarily require the ground truth object pose. In fact, if  measurements are very noisy,  the computed performance index might be low even if it is associated to local minima and corresponds to a completely wrong localization (Fig \ref{locmin}).
 
 \begin{figure}
 	\centering{
 		\includegraphics[scale=0.3]{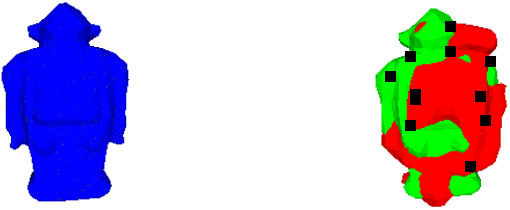}}
 	\caption{On the left: a robot toy in the real pose. On the right: two different estimated poses, both featured by a performance index of 0.008$[m]$ with respect to the set of measurements, coloured in black. The green one corresponds to the correct pose, whereas the red one is a local minimum, representing a completely wrong pose, but anyway consistent with  the measurements.\label{locmin}}
 \end{figure}
 \subsection{Simulation results}\label{sime}
 \indent \textcolor{black}{Table \ref{simres} provides, for each considered object, the following metrics averaged over $50$ independent trials of the MUPF: 
standard localization error in both position and orientation,
performance index $\mathcal{I}_L$ defined in (\ref{pi}),
execution time and reliability.
\GiuliaDelete{and different sets of contact point measurements}
Table \ref{meas&m} reports the total number of measurements $L$ and the  MUPF window size $m$ used for each object.} \\
\indent It is worth  underlining how, when an adequate choice of $m$ is adopted (Section \ref{mupfsec}), \GiuliaDelete{the index $\mathcal{I}_L$}\GiuliaAdd{the localization errors} averaged over trials \GiuliaAdd{are small (e.g. the index $\mathcal{I}_L$} is less than  $2 \, [mm]$, see Fig. \ref{simtest}), the execution time is acceptable and the reliability is high. 
\GiuliaDelete{In particular, different tests on the same object and set of measurements but with different values of $m$
show how MUPF
successfully localizes objects even by exploiting a small number of past measurements at each time step. }\GiuliaAdd{In Fig. \ref{mvalue-study-sim}, the  behavior of the performance index
$\mathcal{I}_L$ is shown as a function of the memory $m$ ranging from $1$ to $L$ (the total number of available measurements). Such plots highlight how MUPF is capable of solving the problem even with small $m$ ($1 < m \ll L $) whereas the standard UPF (i.e. MUPF with $m=1$) doe not converge at all. In addition, Fig. \ref{reliab-study} demonstrates that the algorithm is reliable even with small values of $m$ (provided $m > 1$).}
    \begin{figure}
    	\centering{
    		\includegraphics[scale=0.225]{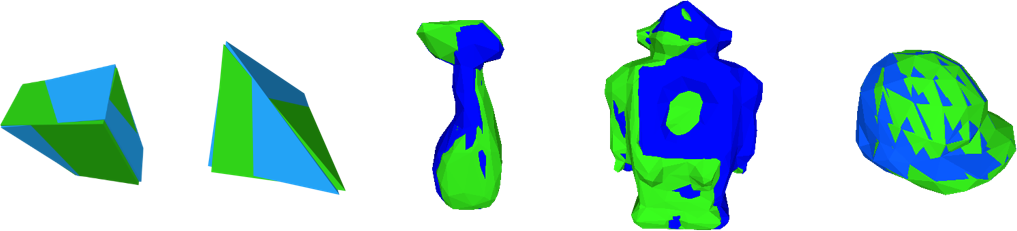}
    	}
    	\caption{MUPF simulation results: the real poses are  coloured in blue, whereas the estimated ones, featured by an error index of 0.002 [m], are coloured in green.\label{simtest}}
    \end{figure}
  \begin{table}[h!]
  {\centering
  \caption{Simulation results for the MUPF algorithm \label{simres}}}
   \begin{tabular*}{\columnwidth}{c  c c c c } 
   \hline
   \hline
    &&&&\\
    Object& \GiuliaAdd{Standard error [deg], [m]} &  $\mathcal{I}_L$ [m]& Time [s]& Succ./Trials\\
  Box & 0.30 - 0.0036 & 0.0025& 1.61 & 50/50\\
  Tetra. & 17.1 - 0.0061 & 0.0021& 3.63 & 50/50\\
  Cleaner& 0.78 - 0.0027 & 0.0025 & 7.32 & 50/50\\
  Robot& 19.6 - 0.0072 & 0.0021 & 3.95 & 50/50\\
  Helmet&  0.06 - 0.0023& 0.0017 & 4.82 & 50/50\\
  
  &  &&&\\ 
          \hline
          \hline
     \end{tabular*}
     \end{table}
       \begin{table}[h!]
       	{\centering
       		\caption{\GiuliaAdd{Simulation Results: measurements and $m$ values} \label{meas&m}}}
       	\centering
       	\begin{tabular*}{0.6\columnwidth}{c c c c c c } 
       		\hline
       		\hline
       		&&&&&\\
       		Object& \GiuliaAdd{$L$} & $m$ &Object& $L$ & $m$\\
       		Box & 15 & 10 &Tetra. & 30 & 15\\
       		Cleaner& 62 & 20 &Robot& 40 & 20\\
       		Helmet& 60 & 30\\
       		
       		&  &&&&\\ 
       		\hline
       		\hline
       	\end{tabular*}
       \end{table}
     
     \begin{figure} 
     	\centering
     	\subfigure[]{
\includegraphics[scale=0.29]{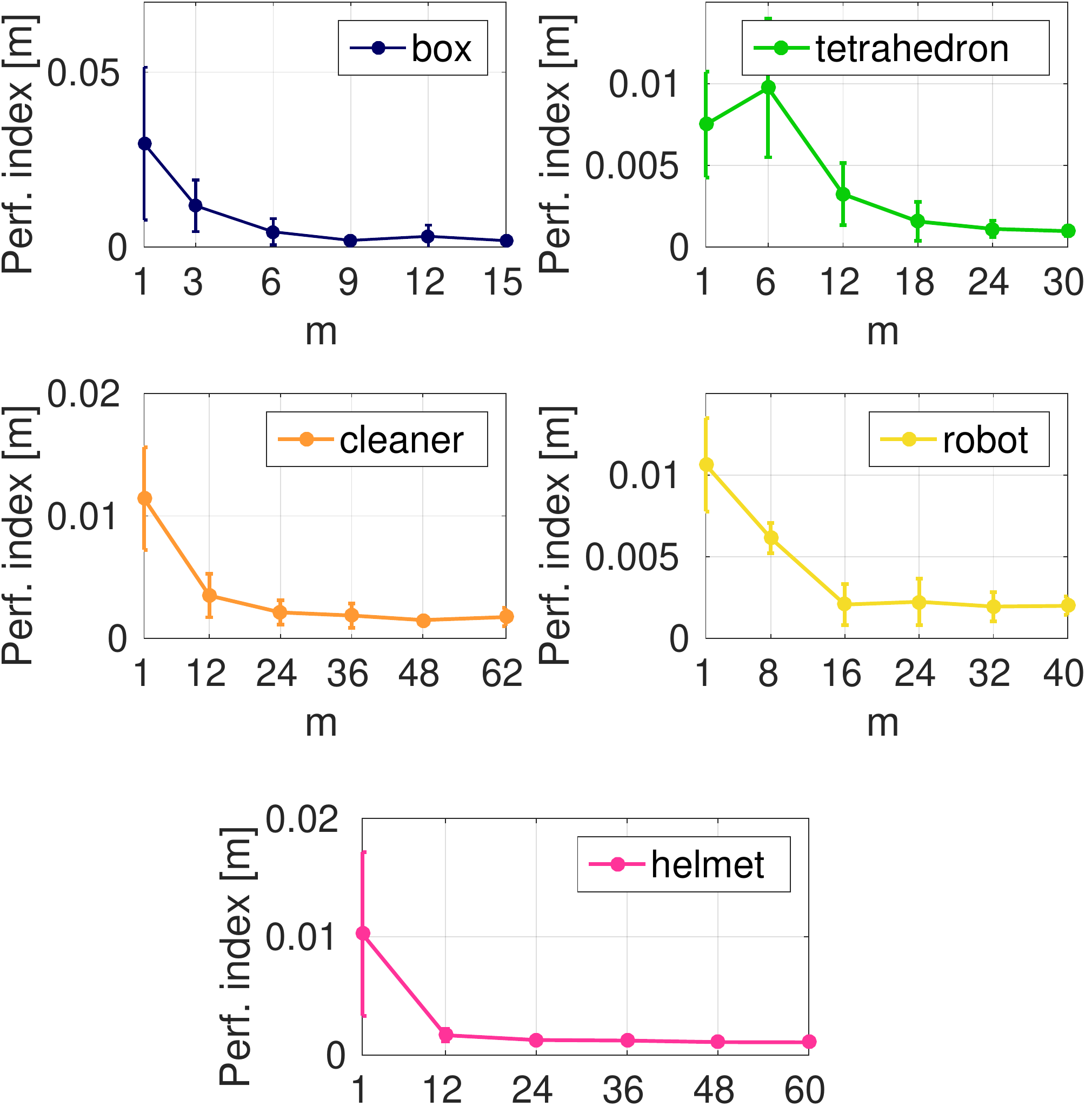}
\label{mvalue-study-sim}}
     \subfigure[]{\includegraphics[scale=0.29]{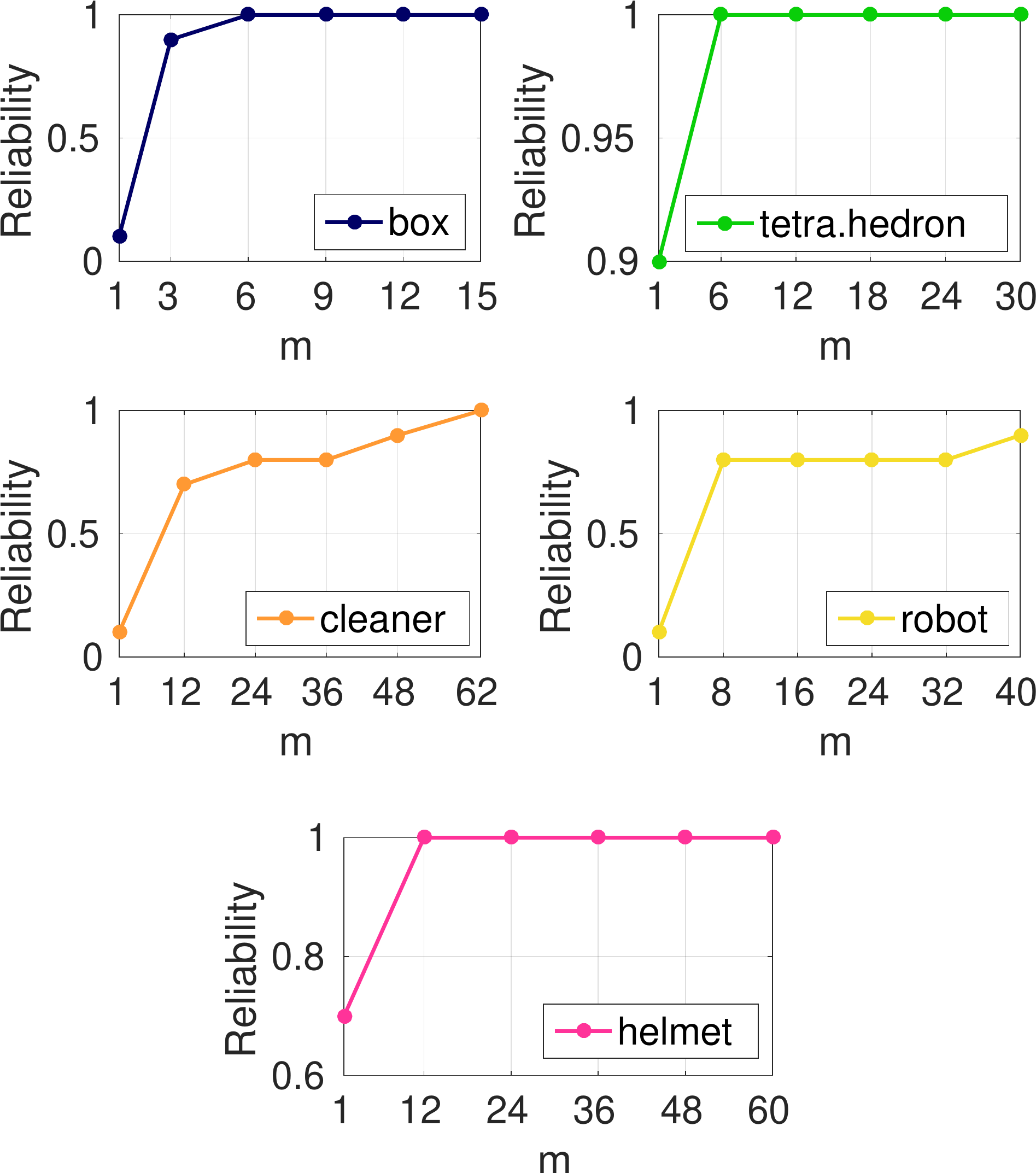}
     	\label{reliab-study}}
     \caption{\GiuliaAdd{MUPF simulation results: (a) average performance index  and (b) reliability (number of successes among trials) on fifty trials by varying $m$, ranging from 1 up to the total number of measurements $L$.}}
     \end{figure}
\indent 
For the sake of comparison, 
\GiuliaAdd{a  simple batch baseline, the \textit{ICP} algorithm~\cite{icp}, and a state-of-art approach,} the \textit{Scaling Series} algorithm presented in \cite{Petr} \GiuliaAdd{specifically for tactile localization}, have been 
applied to the same simulation scenario. \\
\indent \GiuliaAdd{In order to adapt ICP to the tactile localization problem, two point clouds are considered: one consisting of the measurements, and the other representing the object model in the right pose. To this end, suitable models have been obtained by sampling $1000$ points on the object mesh models of Fig. \ref{simobjects}. However, it was found that a standard implementation of ICP does not converge in such a scenario.}\\
\indent \textcolor{black}{The results obtained with the Scaling Series are reported in Table \ref{ss} for the same sets of measurements used in the MUPF simulation tests (Table \ref{meas&m}). For the sake of conciseness, only the values of the performance index $\mathcal{I}_L$ are shown.}
  \begin{table}[h!]
  \centering
  \caption{Simulation results for the Scaling Series algorithm\label{ss}}
  \centering
   \begin{tabular*}{0.9\columnwidth}{c  c  c c } 
   \hline
   \hline
    &&&\\
    Object&  $\mathcal{I}_L$ [m]&  Time - Max. Time [s]& Successes/Trials\\
   Box &  0.001&3.47 - 13.2 &45/50\\
  Tetra.  &0.001 &0.05 - 1.03 &50/50 \\
   Cleaner   & 0.006 & 0.03 - 5.64& 42/50\\
   Robot & 0.003 & 0.02  - 3.64 & 43/50\\
   Helmet & 0.005 &  0.04 - 4.20 & 32/50\\
  &  &&\\
          \hline
          \hline
     \end{tabular*}
     \end{table}

Notice that the execution time of the Scaling Series algorithm significantly changes over the trials as the algorithm generates quite different numbers of particles from trial to trial. 
 Hence, Table~\ref{ss} reports both  \emph{average} and \emph{maximum} (worst-case) execution times.
 Nevertheless, the Scaling Series algorithm proves to be relatively faster than MUPF. 
 In terms of localization precision in the successful trials, the MUPF and Scaling Series algorithms exhibit comparable results. 
 It is worth pointing out, however, that in a non negligible number of trials the Scaling Series algorithm diverged and failed to find a solution. 
 This is somewhat surprising as MUPF has always been executed with the same parameters, whereas the parameters of the Scaling Series algorithm have been specifically tuned to each case in order to achieve better performance. 
 In summary, MUPF proved to be more reliable than the Scaling Series algorithm.

     \section{Algorithm validation with real measurements}\label{exp}
     An extensive evaluation of the MUPF algorithm is performed by tackling the $6$-DOF tactile localization problem for real objects via actual tactile measurements. 
     For these experiments, the  employed code implementation and hardware computing platform are the same ones exploited for the simulation tests.\\ 
     \subsection{Experimental setup}
     \indent Four everyday objects are considered: two toys, the cleaner spray and the robot toy. The experimental tests on the safety helmet are not shown since  many local minima, corresponding to different poses and featured by the same localization error, are wrongly given as possible solutions. The reasons of this behaviour will be  explained in detail in Section \ref{expe}.  The mesh models of the first two objects are reconstructed from ruler measurements (Fig. \ref{obj2}), since they are well-represented by geometrical solid figures. The cleaner spray and robot toy mesh models are the same ones exploited for the simulation tests.
Note that in order to avoid object's slip caused by the robot's movements, each object was strictly fixed to a support
    during measurement collection.

     \begin{figure}
     	\centering
     	\includegraphics[scale=0.27]{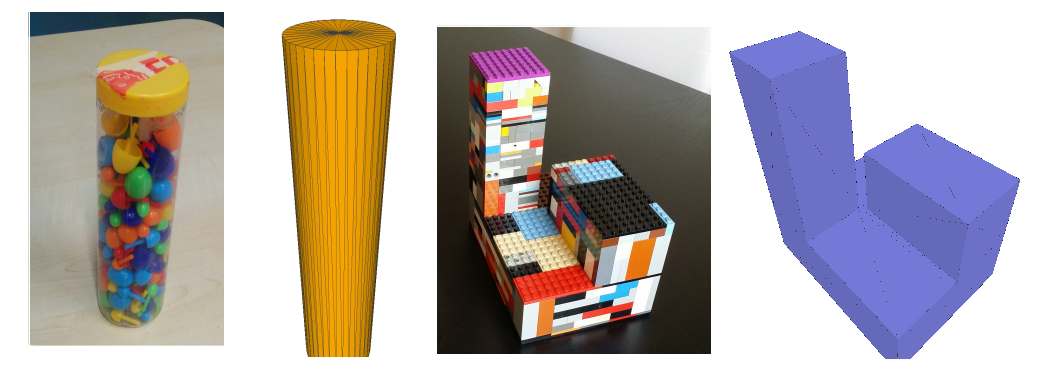}
     	\caption{Mesh models of real   geometric objects. On the left: cylindrical tube, with a diameter of $0.06 \,[m]$ and height of 
			$0.2 \, [m]$,  $144$ triangular faces. 
			On the right: a Lego object, made up of three parallelepipeds (total dimensions of $0.2 \times 0.1 \times 0.2 \, [m^3]$), $36$ triangular faces.\label{obj2}}
     \end{figure}

     \indent The  platform  used for the collection of tactile measurements is iCub, a $53$ degree-of-freedom humanoid robot of the same size as a three
     or four year-old child \cite{icub}. Tactile measurements are supplied by  fingertips on the iCub
      hands, that are covered with capacitive tactile sensors capable of providing accurate contact point measurements, once contact with the object is detected \cite{icub} (Fig. \ref{icub}).
      \begin{figure}

      	\centering{
      		\includegraphics[scale=0.13]{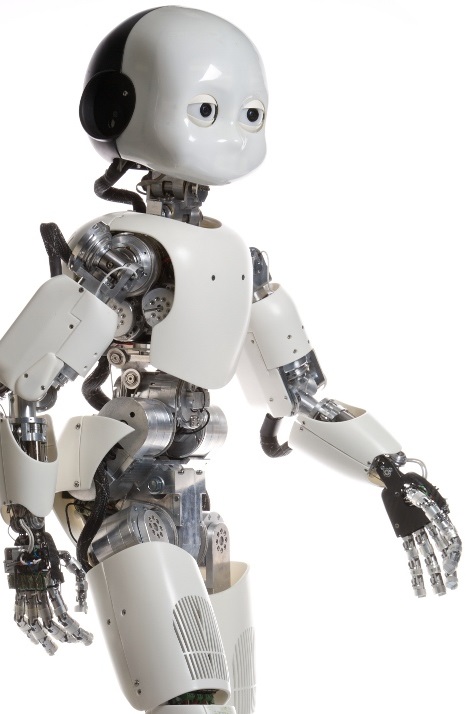}
      	\includegraphics[scale=0.25]{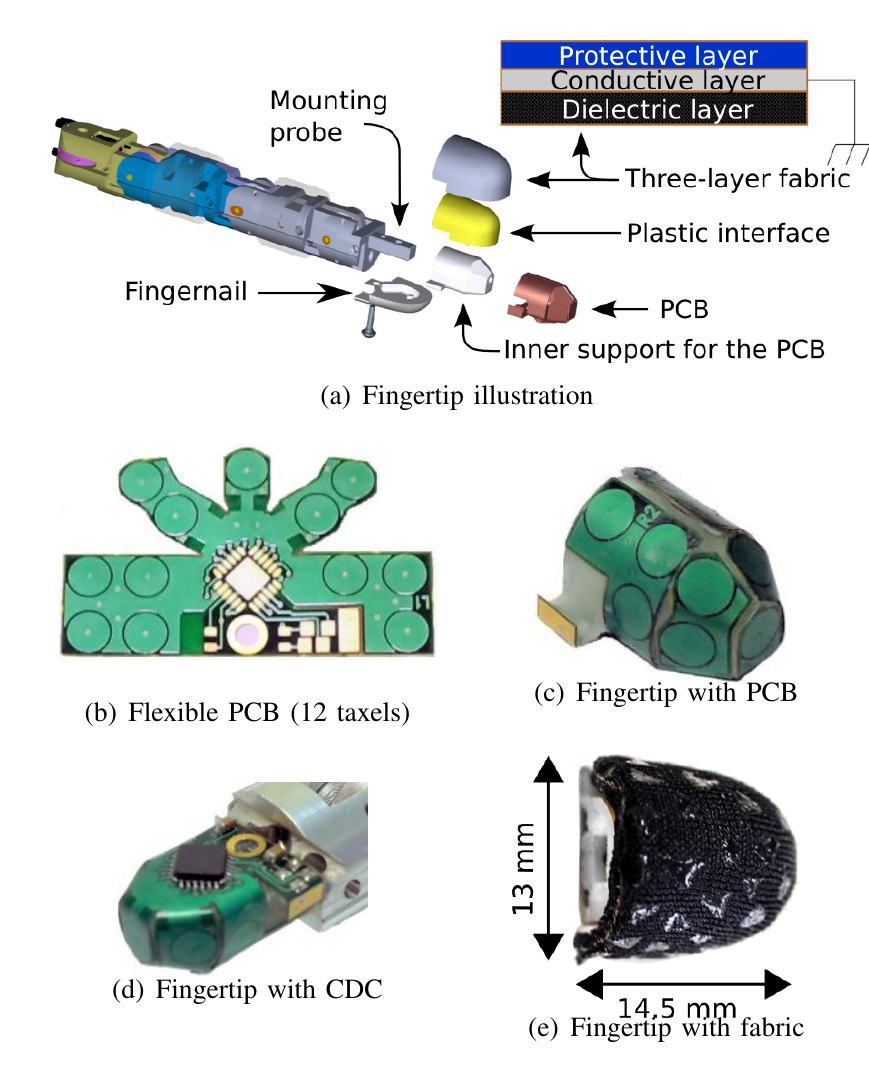}}
      	\caption{On the left: the platform used for the collection of measurements, the humanoid robot iCub. On the right: the iCub fingertip: a)\GiuliaDelete{is a CAD drawing of the 
      		fingertip mounted on the index finger of an iCub robot, b)} $-$ d) show
      		different stages of manufacturing a prototype of the proposed fingertip. The finger consists of multiple layers: the PCB – hosting the
      		CDC converter and the 12 sensors, and a plastic layer,
      		which provides support to the flexible PCB.  The exterior of the fingertip is made of a three-layer fabric:
      		a deformable fabric (dielectric) layer, a conductive layer and a protective
      		layer.\label{icub}}
      \end{figure}
       Due to the object complexity, tactile measurements are collected through a user-guided strategy, consisting of predefined points approximately located around the objects. 
       This strategy was necessary since a completely blind exploration of the objects turned out to be unfeasible and often caused the robot to hit the object with part of the hand not covered with sensors.  
       It is important to remark that, for this work, the final goal of the experimental tests  is the extensive evaluation of the proposed MUPF algorithm through realistic measurements, without focusing on the design of an autonomous measurement collection strategy.

     Before providing experimental results, it is worth discussing the main sources of measurement uncertainty, in order to better appreciate the performance of the proposed algorithm and to understand how to set the  parameters. 
In this respect, one relevant source of uncertainty is given by the tactile sensors themselves. The contact
     point measurement, in fact, is given by the  kinematics of one
     of the  fingers and the supplied $x, y, z$ coordinates are affected by
     calibration offsets. In addition to this, the  kinematics provides
     the $x, y, z$ coordinates of the center of the fingertip. Thus, the retrieved point is always the center of the fingertip even if the tactile
     taxel activation - and thus the contact detection - has taken place on
     the extremity or on the side of the fingertip. Taking into account all these considerations,  tactile measurements were empirically estimated to be affected by  a noise with standard deviation of $0.015 \,  [m]$.  Such  sources of error and uncertainty suggest the values shown in Table \ref{MUPFparex} for the covariance   $\sigma_p$ that characterizes iCub tactile  sensor accuracy.
    \begin{table}[h!]
    	{\centering
    		\caption{Parameter set for the MUPF algorithm\label{MUPFparex}}}
    	\begin{tabular*}{\columnwidth}{c c} 
    		\hline
    		\hline
    		&\\
    		$Q$& diag([10$^{-5}$, 10$^{-5}$, 10$^{-5}$, 10$^{-3}$, 10$^{-3}$, 10$^{-3}$]) [m], [rad]\\
    		$P_0$&  diag([0.04, 0.04, 0.04,  $\pi^2$, $(\pi/2)^2$, $\pi^2$]) [m], [rad]\\
    		$\sigma_p$& 4\text{ }10$^{-4}$[m]\\
    		$\alpha$ & 1  \\
    		$k$ &  2  \\
    		$\beta$ &30  \\
    		$N$ & 1200  \\
    		&\\
    		\hline
    		\hline
    	\end{tabular*}
    \end{table}\\
    \subsection{Experimental results}\label{expe}
    \indent In Tables \ref{expres} and \ref{expresss}, the average performance index, along with the execution time and the algorithm reliability  are provided for 
    fifty trials of both the MUPF and Scaling Series algorithms on the four considered objects.\GiuliaDelete{and different sets of contact point measurements.} \GiuliaAdd{The results obtained with the ICP algorithm are not shown due to the lack of convergence. In addition, only the performance index $\mathcal{I}_L$ is computed in the real experiments, where the true pose is difficult to be measured. Figs. \ref{mvalues-study-exp} and \ref{reliab-exp-study} show the average performance index and the reliability on fifty trials by varying $m$, ranging from 1 (standard UPF) up to the total number of measurements, $m=L$.}
     \begin{table}[h!]
     		\caption{Experimental results for the MUPF\label{expres}}
     		\centering
     	\begin{tabular*}{0.9\columnwidth}{c  c c c c c} 
     		\hline
     		\hline
     		&&&&&\\
     		Object&   $\mathcal{I}_L$ [m]& Time [s]& Successes/Trials & $L$ & $m$\\
     		Lego toy &   0.0090 & 12.8 & 46/50&55&55\\
			Cylinder  & 0.0063 & 6.71 & 50/50&30&18\\
     		Cleaner   & 0.0090 & 13.7 & 50/50&62&30\\
     		Robot   & 0.0054 & 12.3  &43/50&60&36\\
     		&  &&&&\\
     		\hline
     		\hline
     	\end{tabular*}
     \end{table}
      \begin{table}[h!]
      	\centering
      		\caption{Experimental results for the Scaling Series algorithm \label{expresss}}
      	\begin{tabular*}{0.9\columnwidth}{c   c c c  } 
      		\hline
      		\hline
      		&&& \\
      		Object&   $\mathcal{I}_L$ [m]&  Time/ Max. Time [s]& Successes/Trials\\
      		Lego toy  & 0.0073 & 5.03 - 29.71  & 40/50\\
      		Cylinder & 0.0059 & 4.02 - 13.22 & 40/50\\
      		\GiuliaAdd{Cleaner} & 0.0139 & 4.02 - 13.22 & 23/50\\
      		Robot &0.0027& 0.81 - 8.72  & 43/50\\
      		&  &&\\
      		\hline
      		\hline
      	\end{tabular*}
      \end{table}
        \begin{figure} 
        	\centering
        	\subfigure[]{
        		\includegraphics[scale=0.29]{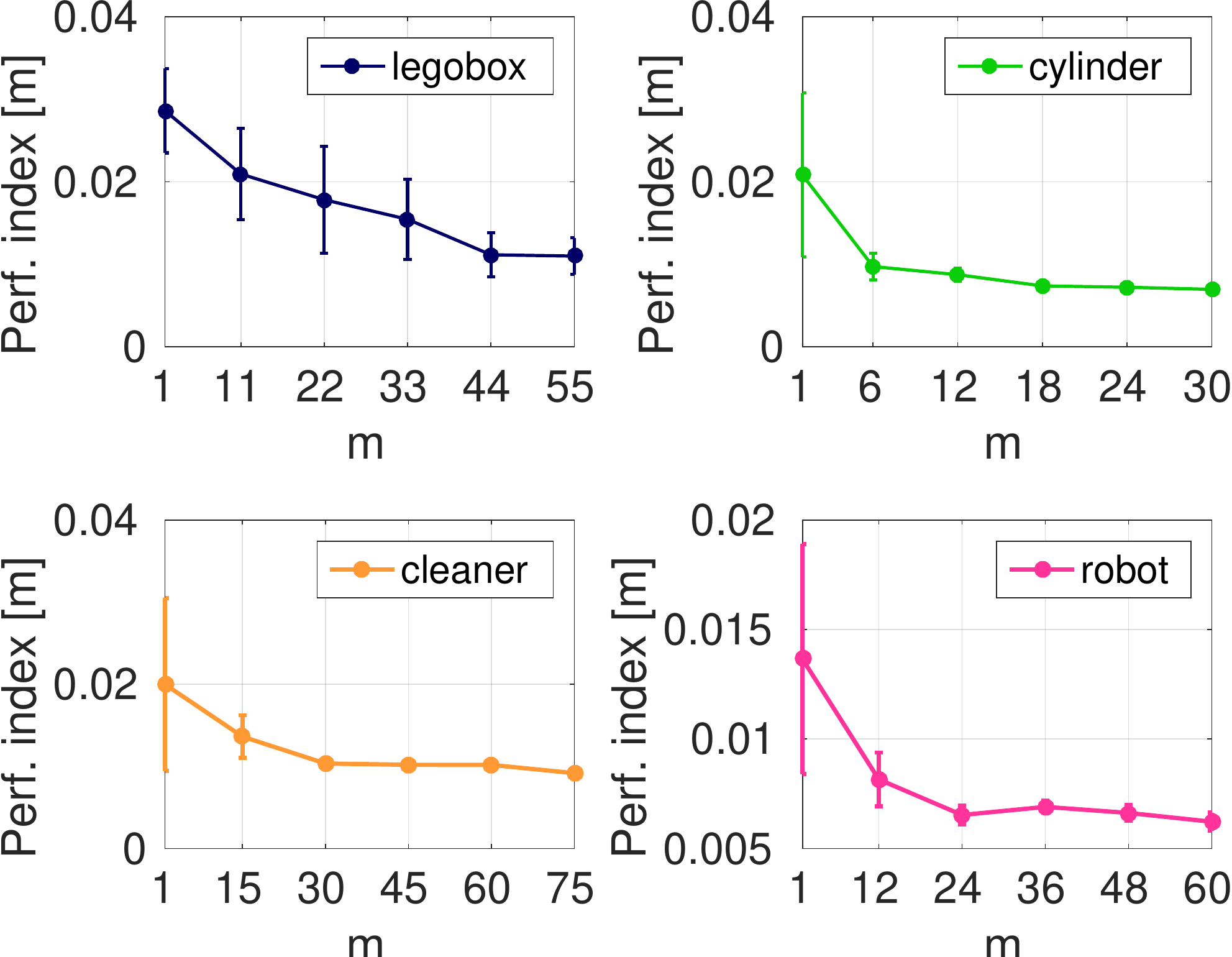}\label{mvalues-study-exp}}
        	\subfigure[]{\includegraphics[scale=0.29]{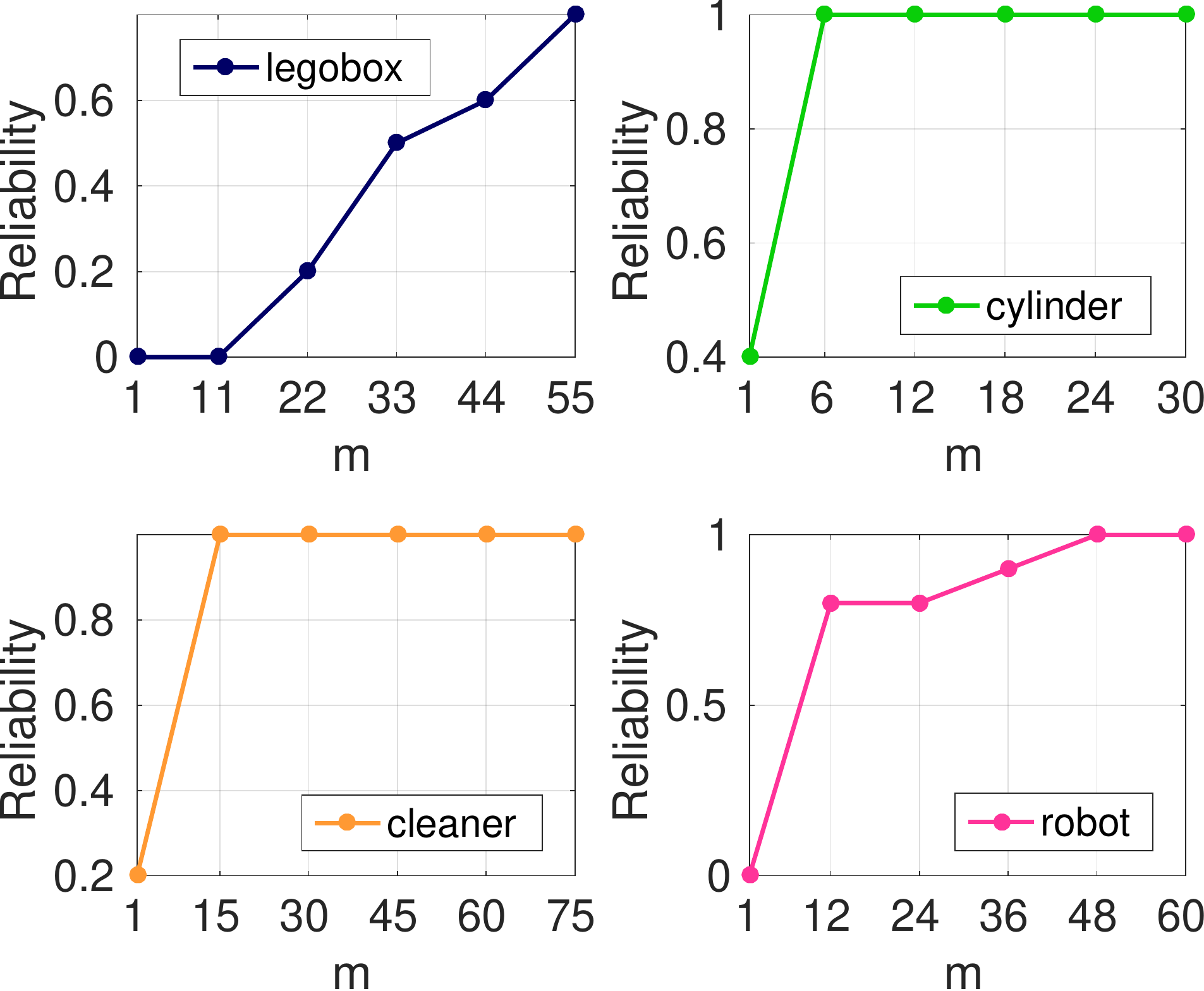}\label{reliab-exp-study}}
        	\caption{\GiuliaAdd{MUPF experimental results: (a) average performance index and (b) reliability (number of successes among trials) on fifty trials by varying $m$, ranging from 1 up to the total number of measurements $L$.}}
        \end{figure}
        
   The experimental tests confirm the MUPF behavior exhibited in the simulation tests,  even if the experimental solutions  are unavoidably affected by a slightly worse performance index, due to the high measurement noise 
		(Fig. \ref{expperf}).  
		The measurement noise is also responsible for the deterioration of algorithm reliability for the Lego and robot toys.  This effect can be ascribed to the fact that the measurement noise is comparable with the dimension of the  distinctive details of these two objects.  In fact, the distinction between a good or a wrong solution is strongly influenced by the object details, since the only exploited information consists of tridimensional points, without taking advantage of surface normals. 
		In such scenarios, a measurement noise of the same entity of the detail dimensions prevents  the user from localizing the object even via visual inspection. 
		As  mentioned above, this is also the reason why experimental tests on the safety helmet are not shown: due to the strongly symmetric shape and the measurement noise, 
		the measurements are not informative enough in the sense that there are many different poses compatible with the measurements (i.e. corresponding to local minima). \\
    \indent  On the contrary, the Scaling Series performance turns out to be much worse compared to what reported in the simulation tests, particularly in terms of reliability.
		The failures of the Scaling Series algorithm are mainly caused by the generation of an insufficient number of particles. 
		Often, it is not simple to set the Scaling Series parameters so that the number of generated particles is sufficient to reliably localize the objects. 
		This shows how parameter tuning  can actually be a weakness of the Scaling Series approach.   
         \begin{figure}
         	\centering{
         		\includegraphics[scale=0.3]{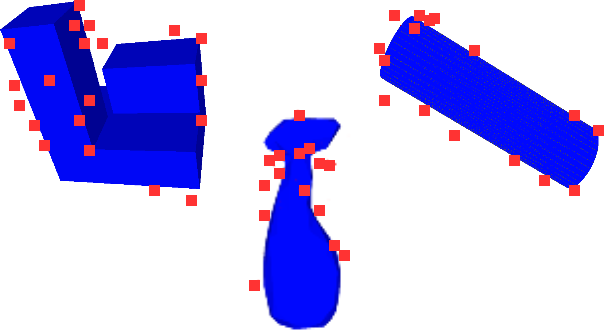}		
         	}
         	\caption{MUPF experimental results: tactile measurements are coloured in red, the estimated poses (performance index of 0.008 [m]) in blue. \label{expperf}}
         \end{figure}    
         \begin{table}[h!]       		      		
         		\caption{Q matrices for algorithm analysis\label{qvalues-sim}}
         		\centering
         	\begin{tabular*}{0.95\columnwidth}{c c} 
         		\hline
         		\hline
         		&\\
         		&Simulated tests\\
         		$Q_1$&diag([10$^{-6}$, 10$^{-6}$, 10$^{-6}$, 10$^{-5}$, 10$^{-5}$, 10$^{-5}$]) [m], [rad]\\
         		$Q_2$&diag(5$\times$\text{}[10$^{-6}$, 10$^{-6}$,10$^{-6}$,10$^{-5}$, 10$^{-5}$,10$^{-5}$]) [m], [rad]\\
         		$Q_3$&diag([10$^{-5}$, 10$^{-5}$, 10$^{-5}$, 10$^{-4}$, 10$^{-4}$, 10$^{-4}$]) [m], [rad]\\
         		$Q_4$&diag(5$\times$\text{}[10$^{-5}$,10$^{-5}$, 10$^{-5}$,10$^{-4}$, $10^{-4}$,$10^{-4}$]) [m], [rad]\\
         		$Q_5$&diag([10$^{-4}$, 10$^{-4}$, 10$^{-4}$, 10$^{-3}$, 10$^{-3}$, 10$^{-3}$]) [m], [rad]\\ 
         		&\\
         		   & Experimental tests\\
         		   $Q_1$&diag([10$^{-6}$, 10$^{-6}$, 10$^{-6}$, 10$^{-4}$, 10$^{-4}$, 10$^{-4}$]) [m], [rad]\\
         		   $Q_2$&diag(5$\times$\text{}[10$^{-6}$, 10$^{-6}$,10$^{-6}$,10$^{-4}$, 10$^{-4}$,10$^{-4}$]) [m], [rad]\\
         		   $Q_3$&diag([10$^{-5}$, 10$^{-5}$, 10$^{-5}$, 10$^{-3}$, 10$^{-3}$, 10$^{-3}$]) [m], [rad]\\
         		   $Q_4$&diag(5$\times$\text{}[10$^{-5}$,10$^{-5}$, 10$^{-5}$,10$^{-3}$, $10^{-3}$,$10^{-3}$]) [m], [rad]\\
         		   $Q_5$&diag([10$^{-4}$, 10$^{-4}$, 10$^{-4}$, 10$^{-2}$, 10$^{-2}$, 10$^{-2}$]) [m], [rad]\\ 
         		   &\\
         		\hline
         		\hline
         	\end{tabular*}
         \end{table}  
         \GiuliaAdd{\subsection{Further analysis} \label{analysis}
         In this section, additional results are provided, with the aim of better analyzing MUPF performance.\\ \indent First, the algorithm robustness has been tested by varying some algorithm parameters, such as the covariance $Q$  of the artificial process noise and  the number of particles $N$. The box-plots of Figs. \ref{q-error} and \ref{q-reliab} point out how the performance index and reliability are not significantly affected by varying the covariance matrix $Q$. 
Fifty trials of the MUPF  have been carried out for five different $Q$ matrices shown in Table \ref{qvalues-sim} (a total of $5 \times 50$ trials). 
The performance index $\mathcal{I}_L$ and reliability averaged over the $50$ trials -  $5$ values for each object - are used in building each box. The box-plots of Fig. \ref{Q}  show the performance obtained with real measurements. 
Due to space considerations, we do not provide plots about the influence of the number of particles $N$ on MUPF performance, since no significant changes have been found by varying $N$ from 
$700$ to $1200$.\\
         \indent Secondly, MUPF execution time has been studied by varying the number of particles $N$ and the MUPF window size 
$m$. 
Figs. \ref{times-m} and \ref{times-N} show the average execution time over fifty trials versus $m$ and, respectively, $N$ in the case of real measurements.
                 \begin{figure} 
                 	\centering
                 	\subfigure[]{
                 		\includegraphics[scale=0.29]{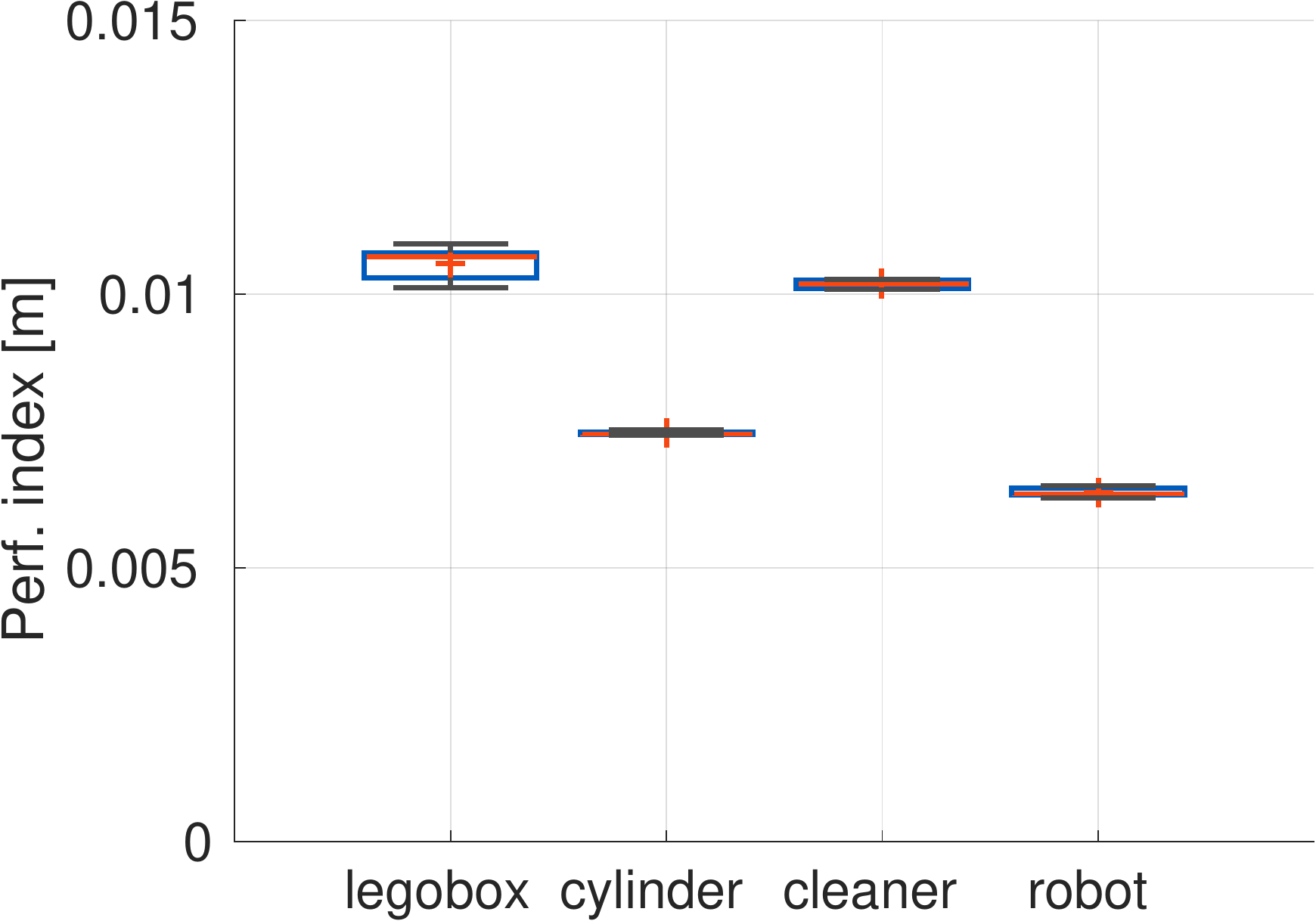}\label{q-error}}
                 	\subfigure[]{\includegraphics[scale=0.29]{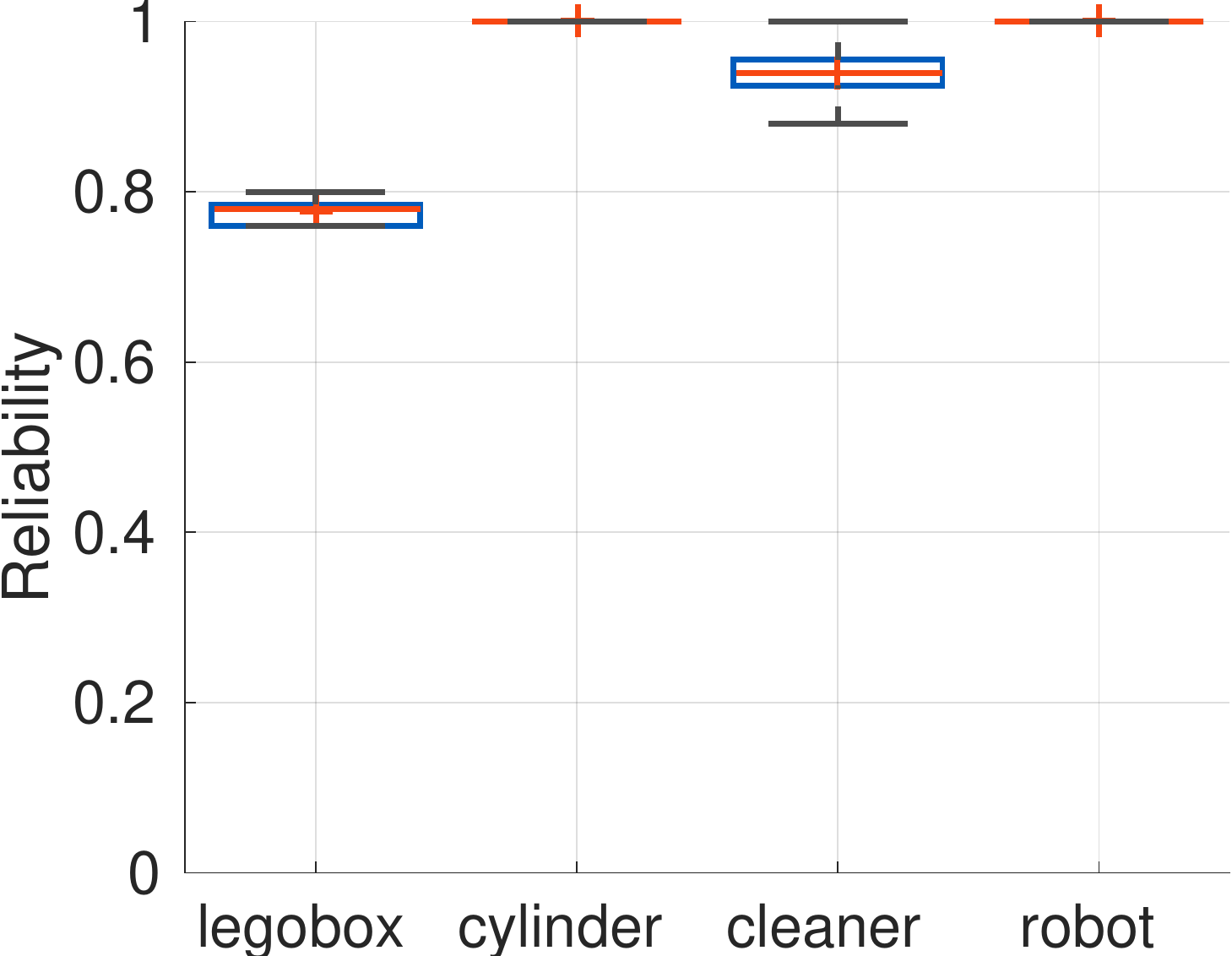}\label{q-reliab}}
                 	\caption{\GiuliaAdd{MUPF robustness analysis: (a) performance index and (b) reliability (number of successes among trials) on fifty trials for 5 different $Q$ values, \GiuliaAdd{shown in Table \ref{qvalues-sim}}.}}
                 	\label{Q}
                 \end{figure}
                 \begin{figure} 
                 	\centering
                 	\subfigure[]{
                 	\includegraphics[scale=0.29]{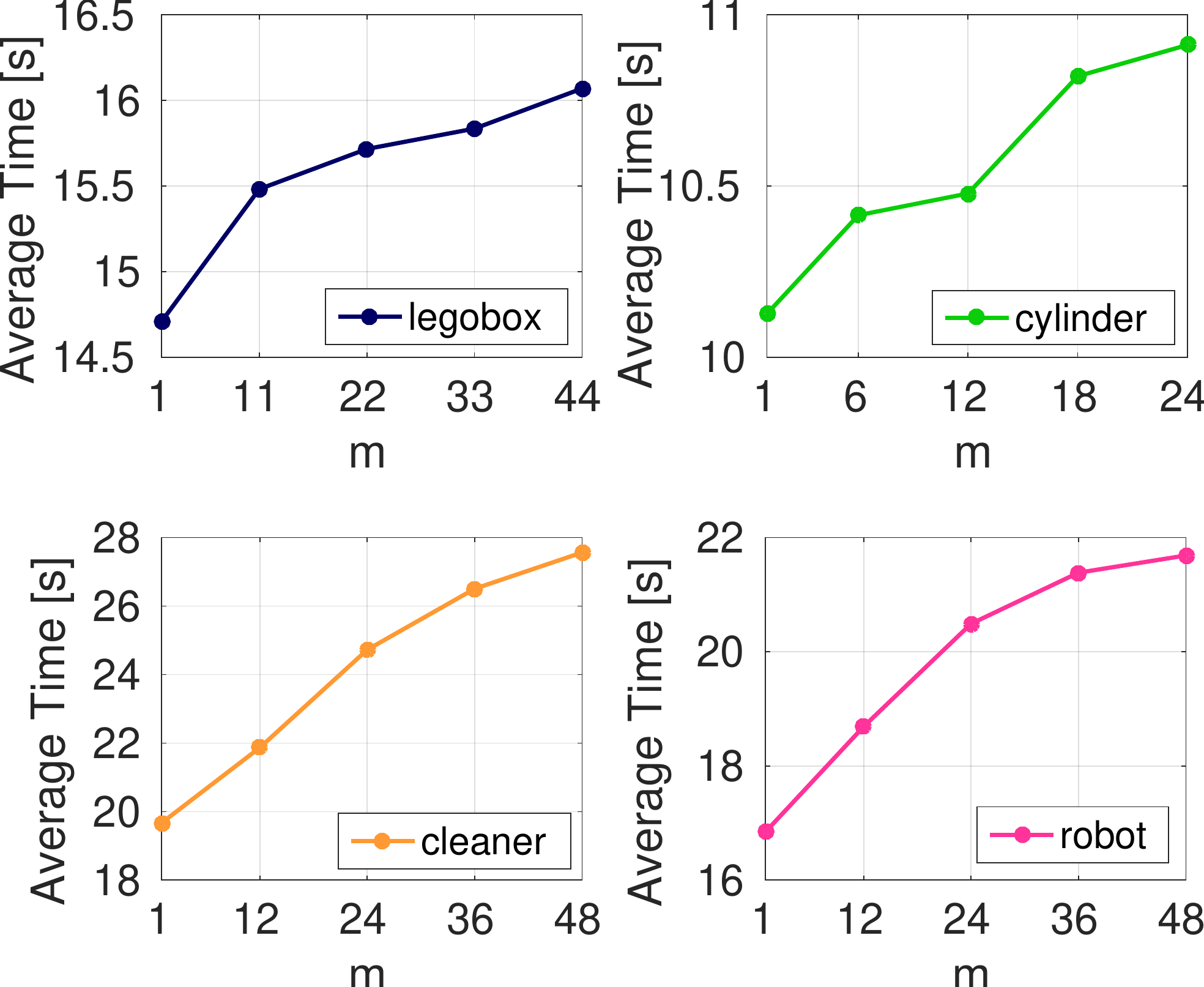}\label{times-m}}
                 	\subfigure[]{\includegraphics[scale=0.33]{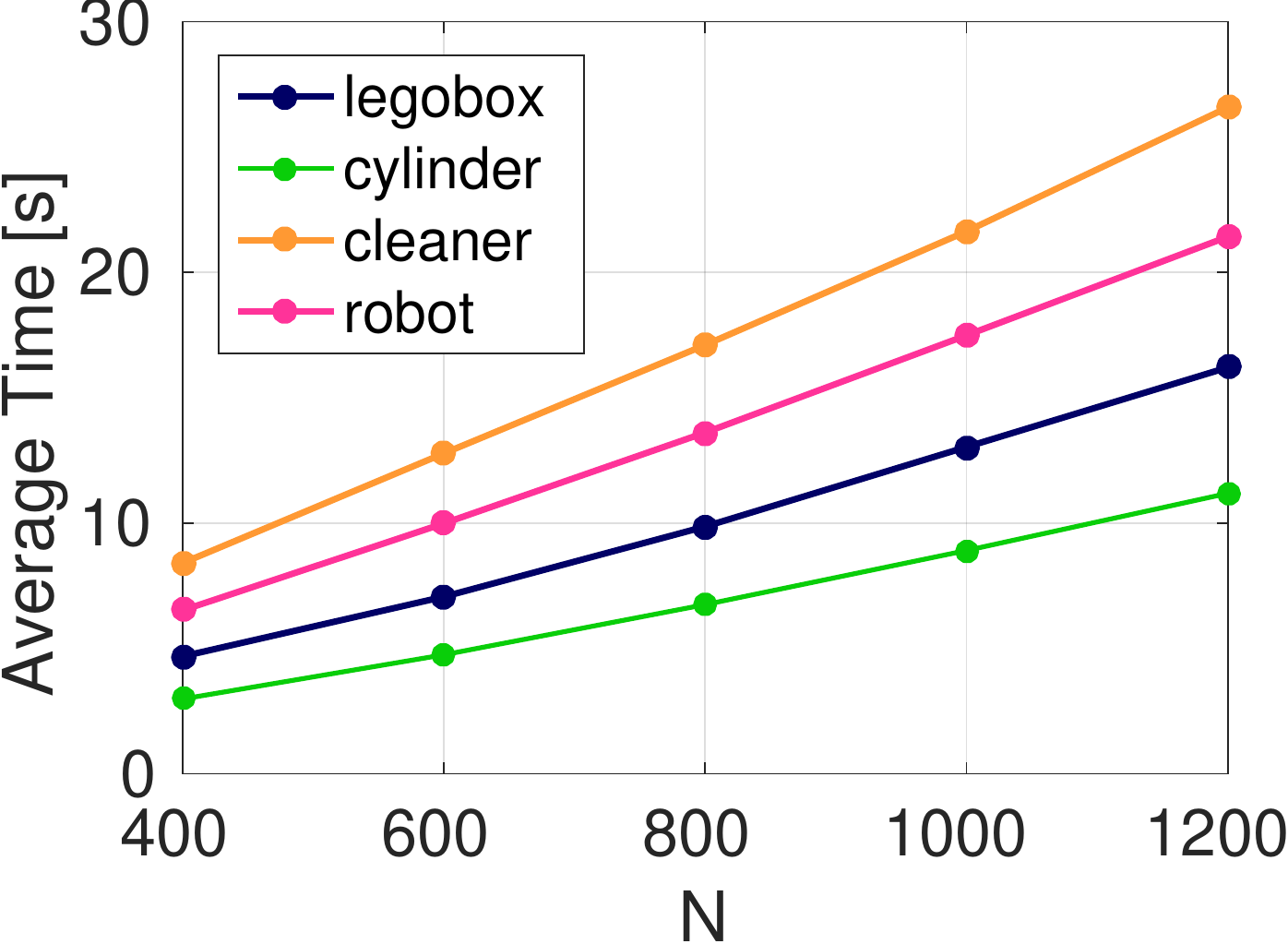}\label{times-N}}
                 	\caption{\GiuliaAdd{Execution time analysis: average execution time on 50 trials a) by varying $m$ values and (b)  by varying $N$.}}
                 \end{figure}\\
         \indent Finally, given the recursive nature of the algorithm, it is worth to analyze the evolution of the performance index 
				$\mathcal{I}_t$ during the MUPF iterations in order to check if it could be used as an appropriate stopping criterion for recursive, on-line localization. 
				 Fig. \ref{locerro} shows how the index $\mathcal{I}_t$ evolves in time, i.e. while new measurements are being processed. 
				It turns out that, after a burn-in period, $\mathcal{I}_t$ quickly converges to a small value. \LorenzoAdd{This suggests that the localization could be terminated whenever the addition of a new measurement (or a sequence of measurements) does not corresponds to a significant reduction of $\mathcal{I}_t$.}
}
            \begin{figure} [h]
            	\centering
           \includegraphics[scale=0.34]{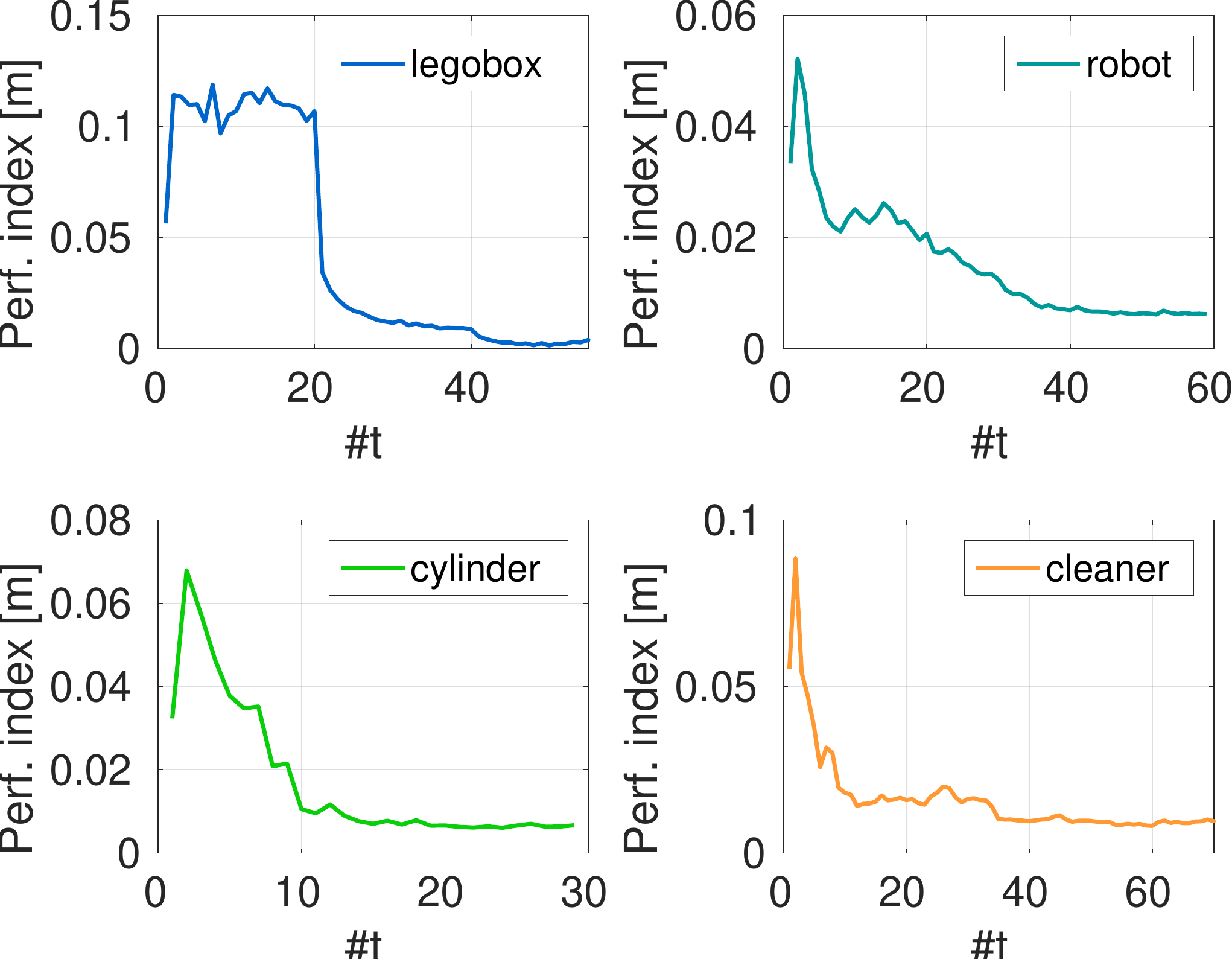}
            	\caption{\GiuliaAdd{Performance index trend at each algorithm time step (with real measurements). After a burn in period,  performance index decreases and converges to a final value.}\label{locerro}}
            \end{figure}     
\section{Discussion}\label{disc}
The proposed solution to the $6$-DOF tactile localization is based on a novel recursive Bayesian estimation algorithm, 
the Memory Unscented Particle Filter (MUPF).  In contrast to  optimization techniques, 
Bayesian filtering turns out to be a successful approach to account for noisy sensors and  inaccurate models. 
A further advantage of the Bayesian approach is that it can be naturally extended to consider the case in which the object moves, 
by introducing a suitable probabilistic model for the object motion.
The multimodal nature of the problem makes particle filtering techniques more suitable for tactile localization than 
nonlinear Kalman filtering approaches. 
However, the exploitation of standard particle filtering for $6$-DOF tactile localization would require a number of particles in the order of $10^6$ which, 
in turn, might entail an unaffordable computational load for real-time operation.

The proposed  MUPF algorithm is capable of localizing tridimensional objects through tactile measurements with  good overall performance and by exploiting a reduced number of  particles (in the order of hundreds). 
The MUPF algorithm relies on the Unscented Particle Filter suitably adapted to the localization problem of
interest.  The Unscented Particle Filter jointly exploits the potentials of the particle filter for approximating multimodal distributions and of the unscented Kalman filter for efficiently generating 
the proposal distribution.
It is worth to point out that, for measurement update purposes, the particle filter requires a probabilistic sensor description in terms of likelihood function while the unscented Kalman filter needs a measurement function allowing to  predict the measurement given the estimated state.
In the specific problem of interest, it is quite natural to characterize the tactile sensor in terms of likelihood (i.e. probability distribution of the sensed contact point given the object pose) while it is clearly not possible
to uniquely predict the sensed contact point given the estimated object pose.
To circumvent this difficulty and be able to apply UPF to tactile localization, the following idea has been pursued:
for given object pose and measured contact point, 
define the likelihood in terms of distance between the object and the measured contact point and
take the predicted contact point as the point on the boundary of the object at minimum distance from the measured contact point.
As a further contribution of this paper, the standard UPF algorithm has been modified by the inclusion of a suitable sliding \textit{memory} (hence the name MUPF) of past measurements in the update of the particle importance weights.
In this respect, it was found that the \textit{memory} feature is crucial for a careful exploitation of the available contact point measurements with consequent improvement of localization accuracy.
 
Furthermore, it is worth underlining how the proposed algorithm succeeds in solving the problem by using only tridimensional contact point measurements, without requiring the knowledge of surface normals.

Performance evaluation, carried out via simulation tests on two geometric objects and three everyday objects by using simulated measurements and tridimensional mesh models reconstructed by vision, demonstrates that the algorithm is reliable and has good performance with an average localization error less than $0.002 \, [m]$ and a computing time of a few seconds. 
Moreover, the algorithm manages to localize real objects with actual tactile measurements  collected with the humanoid robot iCub. The results of experimental tests on four real objects confirm the results of the simulation tests, providing localization errors less than $0.01 [m]$ with a computing time less than $8 \, [s]$.\

The same simulation and experimental tests have been carried out also with a reference algorithm in the literature, called Scaling Series. The obtained results show how the MUPF is competitive with the state of art for $6$-DOF tactile localization, and also exhibits several advantages with respect to the Scaling Series algorithm.

The contributions of this paper suggest several possible perspectives for future work on $6$-DOF object tactile localization. 
First of all, dealing with the localization of objects in presence of slippage is fundamental in real applications. When filtering techniques (e.g. variants of particle filtering) are employed in place of optimization methods, the
extension to this case can be achieved by further considering a suitable model for the object motion. 
Moreover, the nearly recursive nature and the promising computing time of the proposed algorithm would allow reducing localization 
uncertainty on-line during measurement collection. 
In addition, tactile sensors should be assisted by a stereo vision system, both during
the exploration task and for the solution of the localization problem.
In this respect, the measurement model exploited by the algorithms
does not necessarily require tactile measurements: it is sufficient to
have tridimensional points in the space. 
Thus, a possible solution for
a high resolution algorithm for $6$-DOF object localization is to feed the
algorithm with measurements coming both from tactile sensors and,
for example, a stereo vision system, since cameras are often available
on humanoid robots. Finally, a natural extension of the localization problem is the object recognition task. A robot able to localize an object using tactile sensors can also
recognize it among a finite set of possible objects, using the same information.
For example, given \GiuliaDelete{that}\GiuliaAdd{an effective} localization algorithm\GiuliaDelete{ is computationally effective}, the robot could run it\GiuliaDelete{in real-time} with different known
object models and select the one that best matches the observations. \GiuliaAdd{ In \cite{tact-humanoids}, an application of the MUPF algorithm to tactile object recognition is proposed and successfully tested on a challenging set of objects.}

\section{Conclusions}\label{conc}

In this paper, the $6$-DOF tactile localization problem has been efficiently solved by means of a novel recursive Bayesian estimation algorithm, 
the Memory Unscented Particle Filter (MUPF).  The algorithm  is able to estimate in real-time the pose of a tridimensional object by 
only exploiting contact point measurements.  Performance evaluation carried out both via simulation and experimental tests on differently shaped objects has demonstrated the effectiveness of the approach,
also in comparison to the state of the art.


%

\section*{Acknowledgment}
This research has received funding from the European Union\textquoteright s Seventh
Framework Programme for research, technological development and demonstration
under grant agreement No. 610967 (TACMAN).

\ifCLASSOPTIONcaptionsoff
  \newpage
\fi

\bibliography{mupf}

\end{document}